%% file: main.tex
\documentclass{article}

\usepackage[final]{neurips_2023}

\usepackage[utf8]{inputenc} %
\usepackage[T1]{fontenc}    %
\usepackage{hyperref}       %
\hypersetup{
    colorlinks,
    citecolor=blue,
    filecolor=black,
    linkcolor=blue,
    urlcolor=blue
}
\usepackage{url}            %
\usepackage{booktabs}       %
\usepackage{amsfonts}       %
\usepackage{nicefrac}       %
\usepackage{microtype}      %
\usepackage[dvipsnames]{xcolor}

\newcommand{\cb}[1]{\textcolor{Mahogany}{#1}}
\newcommand{\br}[1]{\textcolor{BlueViolet}{#1}}

\input{notations/math_commands.tex}

\usepackage{enumitem}

\usepackage{graphicx}
\usepackage{amsmath}
\usepackage{bbm}

\usepackage{booktabs}
\usepackage{makecell}
\usepackage{caption}
\usepackage{appendix}
\usepackage{cleveref}
\usepackage{subcaption}

\usepackage{csquotes}
\usepackage{listings}

\usepackage{wrapfig}
\usepackage{xspace}
\usepackage{multirow}

\newcommand{\mo}{Moral Permissibility Task\xspace}
\newcommand{\ca}{Causal Judgment Task\xspace}

\title{MoCa: Measuring Human-Language Model Alignment on Causal and Moral Judgment Tasks}

\author{%
Allen Nie$^{1*}$ \quad Yuhui Zhang$^{1}$ \quad Atharva Amdekar$^2$ \quad  \\ 
 \textbf{Chris Piech}$^1$ \quad \textbf{Tatsunori Hashimoto}$^1$ \quad \textbf{Tobias Gerstenberg}$^{3*}$ \\
$^1$Computer Science \quad $^2$ICME \quad $^3$Psychology\\
Stanford University \\
*\texttt{\{anie, gerstenberg\}@stanford.edu}\\
}

\begin{document}

\maketitle

\begin{abstract}

Human commonsense understanding of the physical and social world is organized around intuitive theories. These theories support making causal and moral judgments. When something bad happens, we naturally ask: who did what, and why? A rich literature in cognitive science has studied people's causal and moral intuitions. This work has revealed a number of factors that systematically influence people's judgments, such as the violation of norms and whether the harm is avoidable or inevitable. We collected a dataset of stories from 24 cognitive science papers and developed a system to annotate each story with the factors they investigated. Using this dataset, we test whether large language models (LLMs) make causal and moral judgments about text-based scenarios that align with those of human participants. On the aggregate level, alignment has improved with more recent LLMs. However, using statistical analyses, we find that LLMs weigh the different factors quite differently from human participants. These results show how curated, challenge datasets combined with insights from cognitive science can help us go beyond comparisons based merely on aggregate metrics: we uncover LLMs implicit tendencies and show to what extent these align with human intuitions.

\end{abstract}

\input{sections/intro}

\input{sections/related}

\input{sections/task_v2}
\input{sections/experiment}

\input{sections/discussion}

\input{sections/ethics}
\bibliography{neurips_2022,anthology}
\bibliographystyle{iclr2023_conference}

\clearpage

\setcounter{figure}{0}
\renewcommand{\thefigure}{A\arabic{figure}}
\setcounter{table}{0}  %
\renewcommand{\thetable}{A\arabic{table}}
\appendix

\begin{appendices}

\input{sections/appendix}

\end{appendices}

\end{document}

%% file: notations/math_commands.tex
\usepackage{amsmath,amsfonts,bm}

\def\eqref#1{equation~\ref{#1}}

\def\1{\bm{1}}

\DeclareMathAlphabet{\mathsfit}{\encodingdefault}{\sfdefault}{m}{sl}
\SetMathAlphabet{\mathsfit}{bold}{\encodingdefault}{\sfdefault}{bx}{n}

%% file: sections/intro.tex
\section{Introduction}
\label{intro}

Humans rely on their intuition to understand the world. This intuition helps us to understand not only physical events (e.g., one ball caused the other to move) but also complex social situations (e.g., the collapse of Sam Bankman-Fried's FTX caused unprecedented turmoil in the cryptocurrency market). Given a complex set of events, even with a great amount of ambiguity, we can answer questions such as ``What or who caused it?'' 
Our answers to this question reflect how we intuitively understand events, people, and the world around us~\citep{sloman2015causality,pearl2018book}. 
How do humans handle this complexity?

Cognitive scientists have proposed that we do so by organizing our understanding of the world into intuitive theories \citep{gerstenberg2017theories,wellman1992cognitive}. Accordingly, people have intuitive theories of the physical and social world with which we reason about how objects and agents interact with one another \citep{battaglia2013simulation,ullman2017mind-gam,gerstenberg2021csm,baker2017rational,davis2015commonsense,lake2017building}. 
Concepts related to causality and morality form key ingredients of people's physical and social theories. Given a story, humans can readily make causal and moral judgments about the objects and agents involved in the story. 

Studying these human \underline{\textbf{intuitions}} or systematic \underline{\textbf{tendencies}} when making decisions or judgments is the central focus of psychological experimentations. What are these tendencies? How do they influence our judgment? Over the last several decades, using text-based vignettes, psychologists have disentangled what factors influence people's causal and moral judgments. These factors can be understood as the building blocks of our thought processes for making causal and moral judgments.

Over the last few years, large language models (LLMs) have become increasingly successful in emulating certain aspects of human commonsense reasoning ranging from tasks such as physical reasoning~\citep{tsividis2021human}, visual reasoning~\citep{buch2022revisiting}, moral reasoning \citep{hendrycks2020aligning}, and text comprehension \citep{brown2020language,liu2019roberta,bommasani2021opportunities}. 
Here, we investigate to what extent pretrained LLMs align with human intuitions about the role of objects and agents in text-based scenarios, using judgments about causality and morality as two case studies. We show how carefully constructed text scenarios from cognitive science can help shed light on the extent to which humans and LLMs align in a way that goes beyond mere agreement in aggregate.

Prior work on alignment between LLMs and human intuitions usually collected evaluation datasets in two stages. In the first phase, participants write stories with open-ended instructions. In the second phase, another group of participants labels these participant-generated stories~\citep[e.g.][]{hendrycks2020aligning}. The upside of this approach is the ease of obtaining a large number of examples in a short period of time. The downside of this approach is that the crowd-sourced stories are often not carefully written, and that they lack experimental control. 
Here, we take a different approach. Instead of relying on participant-generated scenarios, we collected two datasets from the existing literature in cognitive science: one on causal judgments and another on moral judgments. 
These scenarios were carefully written by researchers with the intention of systematically manipulating one or a few factors that have been theorized to influence people's judgments. These latent factors naturally bring structures to the otherwise plain text stories, which aim to explain human judgments (see \autoref{fig:story_example}). Using these scenarios, we can design a series of experiments that aim to measure the LLMs' alignment with human intuition and use the scientific framework around human judgments to analyze where the differences occur. We have released the code and dataset here: \url{https://github.com/cicl-stanford/moca}

\paragraph{Our Contributions}

\vspace{-3mm}
\begin{enumerate}[itemsep=0pt, wide=0pt, leftmargin=*, after=\strut,label={(C\arabic*):}]
\item We summarize the main experimental findings of 24 cognitive science papers into factors that have been shown to influence participants' judgments on moral and causal stories (see short version in \autoref{tab:short-def-causal}, full version in \autoref{tab:causal_tag_defs}). From these papers, we create a causal and moral judgment challenge set in which only a few words are changed, yet they lead to big differences in people's judgments. We collected 5150 human responses to our stories and annotated the latent factors with experts (see \autoref{fig:story_example}).
\item We evaluate how a wide range of LLMs align with human judgments. This includes models of different sizes, models fine-tuned via reinforcement learning using human feedback (\textbf{RLHF}), and those distilled from instruction-following, like \textbf{Alpaca-7B}. Our results show that larger models and RLHF techniques lead to better alignment.
\item We compute the Average Marginal Component Effect (AMCE) to reveal the implicit tendencies for each factor from human and LLM judgments. 
Our analysis reveals a number of implicit tendency differences between models trained within the same company or trained with the same technique. Our analysis highlights the importance of using a carefully constructed dataset to understand tendency alignment.
\end{enumerate}
\vspace{-3mm}

%% file: sections/related.tex
\section{Related Work}
\vspace{-3mm}

For causal reasoning, there is an active line of research at the intersection of natural language processing (NLP) and commonsense reasoning that involves extracting and representing causal relationships among entities in text. In some cases, these relationships are based on commonsense knowledge of how objects behave in the world \citep{sap2019atomic, bosselut-etal-2019-comet, talmor-etal-2019-commonsenseqa}. In other cases, models identify scenario-specific causal relations and outcomes. Sometimes, the causal relationship between entities is explicitly stated (e.g., \textit{those cancers were caused by radiation}) \citep{hendrickx-etal-2010-semeval}, while at other times the relationship is left implicit and needs to be inferred \citep{mostafazadeh-etal-2016-corpus}. Causal reasoning is also included in broad benchmarks for language understanding, such as the Choice of Plausible Alternatives (COPA) in SuperGLUE \citep{roemmele2011choice,wang2019superglue}. 

For moral reasoning, tasks have focused on evaluations of agents in narrative-like text. These tasks and datasets vary in the amount of structure they provide, ranging from pairs of free-form anecdotes and judgment labels \citep{lourie2021scruples,hendrycks2020aligning}, to inputs with components separated out into norms, intention, actions, and consequences \citep{emelin-etal-2021-moral}. Prior work has also looked into moral dilemmas, particularly around everyday scenarios \citep{hendrycks2020aligning,jiang2021delphi}. In this work, the moral scenarios were generated through crowd-sourcing \citep{emelin-etal-2021-moral,ziems2022moral}, and there is general agreement about what's the morally right thing to do in a scenario (i.e. these scenarios weren't moral dilemmas). Alternatively, scenarios have been scraped from online communities such as Reddit~\citep{forbes-etal-2020-social,lourie2021scruples, roemmele2011choice}. While these scenarios have greater external validity, they lack the experimental control of scenarios from cognitive science research. How LLMs align with people's intuitions across a range of moral dilemmas hasn't systematically been studied before. \cite{kiciman2023causal} proposed modularized tests to understand whether LLMs can perform different types of causal analysis. Our work investigates whether these LLMs exhibit implicit tendencies similar to those of humans. \cite{scherrer2023evaluating} evaluated LLMs on moral scenarios but focused on moral rule violation along the axis of moral ``ambiguity''. There are also discussions of moral behaviors with reinforcement learning agents in game-like environments~\citep{hendrycks2021would,reinecke2023puzzle,pan2023rewards} and with respect to legal reasoning~\citep{almeida2023exploring}.

\begin{figure}[tbp]
  \centering
  \includegraphics[width=\linewidth]{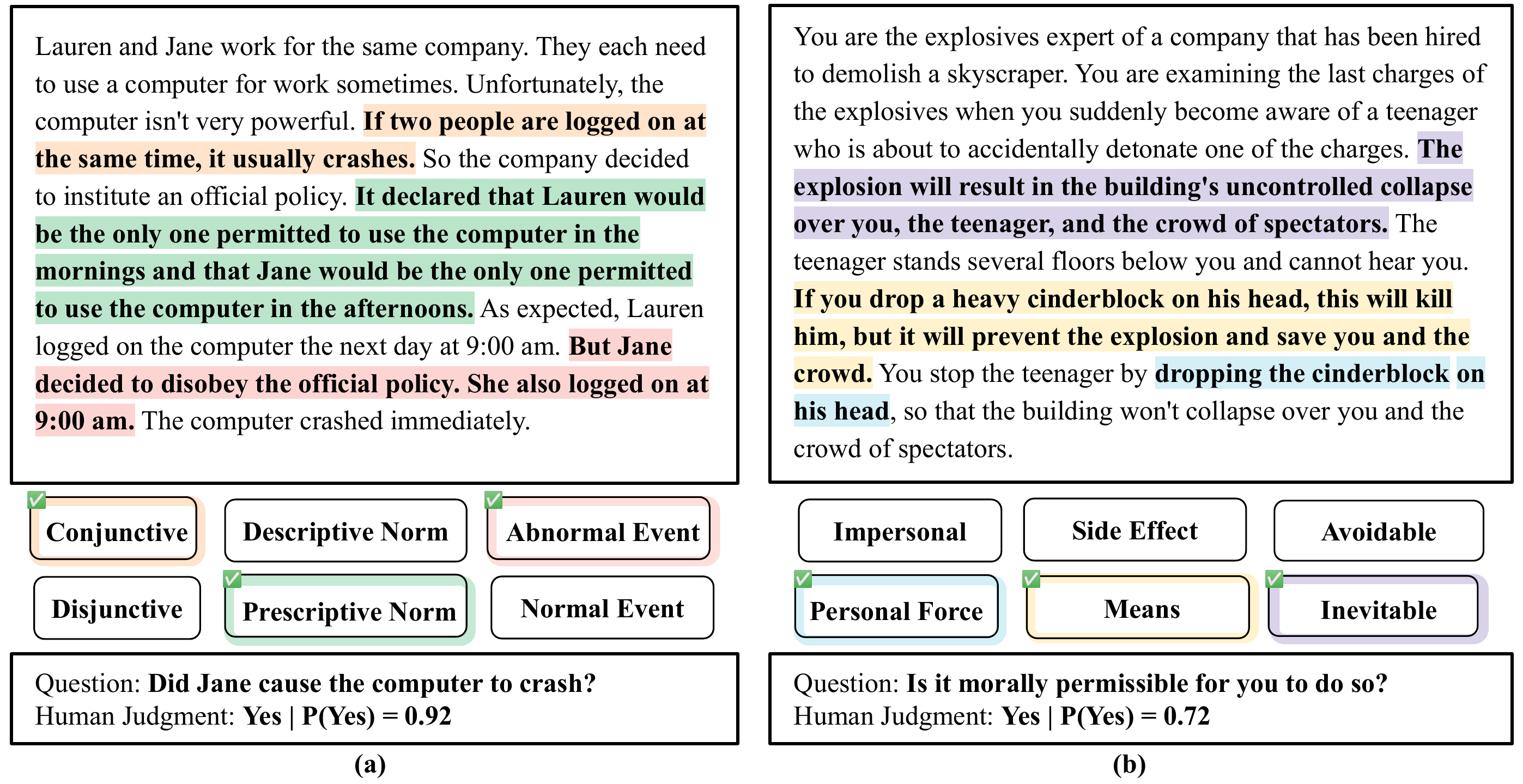}
  \caption{Two examples from our collected dataset. (a) shows a causal judgment story, and (b) shows a moral judgment story. In (a), a conjunction of two events was required, an abnormal event occurred, and Jane violated a prescriptive norm~\citep[scenario taken from][]{knobe2008causal}. In (b), the teenager's death was inevitable; his death is a necessary means to save others, and bringing about his death requires the use of personal force \citep[scenario taken from][]{christensen2014moral}.}
  \label{fig:story_example}
 \vspace{-5mm}
\end{figure}

Our work falls under a broad range of research on commonsense reasoning, where models are encouraged to produce outputs that match human intuitions \citep{gabriel2020artificial,kenton2021alignment}. In some cases, such as physical commonsense reasoning, alignment with human behavior is straightforward, while in the case of social commonsense, the subjectivity and diversity of human intuitions make alignment more challenging \citep{davani2022dealing}. Prior analysis of language models' reasoning abilities includes measuring their behavior in zero-shot settings, after task fine-tuning \citep{jiang2021delphi,hendrycks2020aligning}, or with human-curated support, such as chain-of-thought prompt engineering~\citep{wei2022chain,wang2022self}. Evaluating LLMs with causal hypotheses has also been explored by~\cite{kosoy2022towards}. While they focused on the influence of causal structure, we investigate a variety of factors (including causal structure). Concurrently, \cite{jin2022make} curated a challenge dataset to examine and verify LLMs' ability to conform to three categories of social norms. We look at five factors that have been shown to influence people's moral judgments and analyze whether LLMs respond to these factors similarly to humans.

%% file: sections/task_v2.tex
\vspace{-3mm}
\section{Judgment Tasks}
\vspace{-3mm}

We study the alignment between LLMs and human participants in two case studies: (1) a causal judgment task about whether someone or something caused the outcome of interest, and (2) a moral judgment task where the question is whether what a protagonist did (or failed to do) was morally permissible. \autoref{fig:story_example} shows an example of each task.  

\vspace{-3mm}
\subsection{Causal Judgment Task}

Deciding what ``the'' cause of an outcome was can be challenging because there are often multiple events that contributed to an outcome. Here is an example story:   

\textit{A merry-go-round horse can only support one child. Suzy is on the horse. Billy is not allowed to get on, but he climbs on anyway. The horse broke due to the two children's weight. Did Billy cause the horse to break?}

Even though both children were involved in breaking the horse, when participants are asked to assess whether Billy (who wasn't allowed to get on) caused it to break, they tend to answer with ``yes''. 
Formal models of causal selection have been developed that capture choice patterns in scenarios like this one \citep{kominsky2015causal,icard2017normality,gerstenberg2020physical,quillien2023counterfactuals}. In this example, Billy violated a prescriptive norm, whereas Suzy didn't. People often select norm-violating events as the cause of an outcome \citep{knobe2008causal,hitchcock2009cause,alicke2011causation,hilton1986knowledge,hart1985causation}.
It is worth noting that causality is closely related to responsibility and blame. In fact, formal models of responsibility and blame have been built on top of a causal selection model~\citep{halpern2015cause,lagnado2013causal}. 

In addition to norm violations, prior work has identified a number of factors that systematically influence people's causal judgments. Here, we focus on the following six factors: causal structure, agent awareness, norm type, event normality, action/omission, and time. \autoref{tab:short-def-causal} provides brief definitions of the different factors.

\vspace{-3mm}
\subsection{Moral Permissibility Task}

Philosophers and cognitive scientists have used moral dilemmas to develop and evaluate normative theories of ethical behavior, and to investigate what moral intuitions people have. A popular moral dilemma is the trolley problem \citep{foot1967problem,thomson1976killing}, where the question is whether it's morally permissible for the protagonist to re-route a trolley from the main track (saving some) to a side track (killing others). Some LLMs already show a certain degree of alignment with human responses in various versions of the trolley dilemma~\citep{hendrycks2020aligning,jiang2021delphi}. However, only a few factors varied in these trolley dilemmas, such as the number of people on each track or personal attributes (age, social status, etc.; \citealp{awad2018moral}).

A large number of factors have been shown to influence people's moral intuitions. For example, in \autoref{fig:story_example}b, people may consider factors such as whether personal force was required to bring about the effect~\citep{greene2009pushing}, whether there was a risk for the protagonist themselves \citep{bloomfield2007morality}, whether the harm was a side effect or a means for bringing about the less bad outcome \citep{moore2008shalt}, and whether the harm was inevitable~\citep{hauser2006moral,mikhail2007universal}. This work has shown that people's moral judgments are sensitive to a variety of factors that haven't been considered in existing work on the alignment between LLMs and people's moral intuitions \citep{waldmann2007throwing,liao2012putting,christensen2014moral,kleiman2015inference}. \autoref{tab:short-def-moral} provides brief definitions of these factors. 

\begin{table}[bp]
\centering 
\small
\caption{\textbf{Dataset}: We report dataset statistics on the label distribution, average length of each story, and inter-rater agreement between two annotators on the factors and the sentences they highlight. Additionally, we collect a binary response for each story from 25 people.}
\label{tab:dataset}
\begin{tabular}{@{}rcccccc@{}}
\toprule
Dataset & \# Stories & Yes ($p>0.6$) & No ($p<0.4$) & Ambiguous & \begin{tabular}[c]{@{}c@{}}\# words per \\ story\end{tabular} & \begin{tabular}[c]{@{}c@{}}\# words per \\ translated story\end{tabular} \\ \midrule
Causal  & 144        & 48                       & 50                   & 46        & 162                                                           & 82.9                                                                     \\
Moral   & 62         & 23                       & 10                   & 29        & 72.5                                                          & 53.5                                                                     \\ \midrule
Total   & 206        & 71                       & 60                   & 75        & 135                                                           & 74.1                                                                     \\ \bottomrule
\end{tabular}
\end{table}

\vspace{-3mm}
\subsection{Dataset}
\vspace{-3mm}

For each task, we transcribe stories from a number of papers. For the \ca, we transcribed 144 stories from a selection of 20 papers, and for the \mo, we transcribed 62 stories from 4 papers that covered a wide range of moral dilemmas. High-level descriptions of the dataset are presented in \autoref{tab:dataset}. We select these papers relying on expert domain knowledge -- as these papers contain robust scientific findings and cover different kinds of scenarios.
We collect 25 yes/no answers for each story from a crowd-sourcing platform with IRB approval for our data collection process. We describe the experimental design and data collection details in Appendix~\ref{sec:app:crowd-source}.
We additionally recruited two domain experts to annotate what factors are present in each story.
We report the inter-rater agreement between the two annotators and describe their annotation process in Appendix~\ref{sec:app:inter-rator}.

\begin{table}[t]
\vspace{-3mm}
\centering 
\caption{\textbf{Factors}: Factors that influence causal selection judgments (top) and moral permissibility judgments (bottom). We provide definitions for each factor in Appendix~\ref{sec:app:causal_factors} and Appendix~\ref{sec:app:moral_factors}. See the full version in \autoref{tab:causal_tag_defs}.}
\label{tab:short-def-factors}
\begin{subtable}[t]{0.5\textwidth}
\small
\centering 
\caption{\textbf{Causal Factors}}
\label{tab:short-def-causal}
\begin{tabular}{@{}ll@{}}
\toprule
\textbf{Factor}            & \textbf{Attribute}                  \\ \midrule
\textbf{Causal Structure}  & Conjunctive | Disjunctive       \\ \midrule
\textbf{Agent Awareness}   & Aware | Unaware                 \\ \midrule
\textbf{Norm Type}         & Prescriptive | Statistical      \\ \midrule
\textbf{Event Normality}   & Normal | Abnormal               \\ \midrule
\textbf{Action / Omission} & Action | Omission               \\ \midrule
\textbf{Time}              & Early | Late | Same Time        \\ \bottomrule
\end{tabular}
\end{subtable}%
\begin{subtable}[t]{0.5\textwidth}
\small
\centering 
\caption{\textbf{Moral Factors}}
\label{tab:short-def-moral}
\centering 
\begin{tabular}{@{}ll@{}}
\toprule
\textbf{Factor}            & \textbf{Attribute}                  \\ \midrule
\textbf{Causal Role}  & Means | Side Effect       \\ \midrule
\begin{tabular}[t]{@{}l@{}} \textbf{Locus of Intervention}\end{tabular}  & Instrument | Patient of Harm       \\ \midrule
\textbf{Personal Force}   & Personal | Impersonal             \\ \midrule
\begin{tabular}[t]{@{}l@{}} \textbf{Counterfactual} \\ \textbf{Evitability}\end{tabular}         & Avoidable | Inevitable \\ \midrule
\textbf{Beneficence}   & Benefit Self | Benefit Other        \\ \bottomrule
\end{tabular}
\end{subtable}
\end{table}

%% file: sections/experiment.tex
\vspace{-3mm}
\section{Experiment and Result}
\vspace{-3mm}

\subsection{Do LLMs make the same judgments as people on these stories?}

\paragraph{Setup.} We first directly compare the responses of a set of LLMs to those of human participants across the set of stories in the original cognitive science experiments. We choose a wide range of language models that have achieved good performance with fine-tuning on other natural language understanding tasks ~\citep{radford2019language,devlin-etal-2019-bert,liu-etal-2019-robust,lan2019albert,clark2020electra}. 
We conduct a 3-class comparison to compute accuracy: a response of ``Yes'', ``No', and an additional response of ``ambiguous'' when the agreement between the average human responses or the probability of model output is 50\% ± 10\%. We normalize the model output probability to compute $P(\text{``Yes''}|\text{Story + Prompt})$ and $P(\text{``No''}|\text{Story + Prompt})$. For chat-based APIs, we prompt the model to output these labels directly. See Appendix~\ref{sec:app:r1_exp_details} for details about how we computed the probability for each model. Results should not be compared between completion and chat models.

We report discrete agreement (Agg) between LLMs and human participants. The agreement is computed in the same way as accuracy, but since we don't want to imply that higher agreement is necessarily better, we use the term ``agreement'' instead of ``accuracy''.
We also report additional measures of agreement, such as the area under the receiver operating characteristic curve (AUC), the absolute-mean-squared error (MAE), and cross-entropy (CE) on the probability of the matched label.

\begin{table}[t]
\small
\centering
\caption{\textbf{Original Story}: We ran our experiments and compute the 95\% bootstrapped confidence interval for the result. We report discrete agreement (Agg), the area under the curve for the unambiguous stories (AUC, higher is better matched, 0.5 is chance, and below 0.5 is worse matched than chance), mean absolute error (MAE, lower is better matched), and cross-entropy (CE, lower is better matched). In \autoref{tab:prompt_methods}, we only report results from the best model. \cb{Red} shows the alignment decreased compared to before. \br{Blue} shows the alignment increased. See \autoref{tab:full_social_sim} for a full comparison.}
\label{tab:original_story_exp}
\begin{subtable}[t]{\textwidth}
\caption{Zero-shot with Different Large Language Models}
\label{tab:original_story_zeroshot}
\setlength\tabcolsep{2.7pt}
        \centering
         \resizebox{0.85\textwidth}{!}{
\begin{tabular}{@{}r|cccc|cccc@{}}
\toprule
\multicolumn{1}{l}{}           & \multicolumn{4}{c}{\textbf{Causal Judgment}} & \multicolumn{4}{c}{\textbf{Moral Permissibility}} \\ \midrule
Models                         & Agg ($\uparrow$)              & AUC ($\uparrow$)  & MAE ($\downarrow$) & CE ($\downarrow$)   & Agg ($\uparrow$)              & AUC ($\uparrow$)  & MAE ($\downarrow$) & CE ($\downarrow$) \\ \midrule 
RoBERTa-large           & 34.0{\tiny $\pm$5.6} & 0.50   & 0.50  & 2.6  & 26.6{\tiny$\pm$7.3} & 0.50   & 0.50  & 2.65 \\
ALBERT-xxlarge          & 35.1{\tiny $\pm$5.6} & 0.51  & 0.38 & 1.37 & 23.4{\tiny$\pm$8.1} & 0.48  & 0.37 & 1.82 \\
Electra-gen-large & 34.7{\tiny$\pm$5.9} & 0.53  & 0.44 & 1.94 & 26.6{\tiny$\pm$8.1} & 0.49  & 0.46 & 1.59 \\
GPT2-XL                 & 34.4{\tiny$\pm$5.6} & 0.49  & 0.42 & 1.47 & 26.6{\tiny$\pm$8.1} & 0.51  & 0.43 & 1.71 \\ \midrule 
GPT3-babbage-v1             & 31.2{\tiny$\pm$5.9} & 0.47  & 0.33 & 0.74 & 18.5{\tiny$\pm$7.3} & 0.43  & 0.41 & 1.03 \\
GPT3-curie-v1              & 36.8{\tiny$\pm$6.2} & 0.51  & 0.37 & 1.14 & 26.6{\tiny$\pm$8.1} & 0.45  & 0.39 & 1.32 \\
GPT3.5-davinci-v2             & 37.8{\tiny$\pm$5.6} & 0.61  & 0.31 & 0.72 & 32.3{\tiny$\pm$8.1} & 0.67  & \textbf{0.30}  &\textbf{ 0.74} \\ 
GPT3.5-davinci-v3 & 41.3{\tiny$\pm$5.6} & \textbf{0.67} &  0.41 & 1.39 & 20.2{\tiny$\pm$7.3} & 0.61 & 0.51 & 2.61\\
\midrule 
Alpaca-7B & 32.3{\tiny$\pm$5.6} & 0.51 & \textbf{0.28} & \textbf{0.65} & 32.3{\tiny$\pm$8.1} & 0.45 & 0.32 & 0.91 \\
Anthropic-claude-v1 & 36.1{\tiny$\pm$5.9} & 0.57 & 0.34 & 1.37 & 37.1{\tiny$\pm$8.9} & 0.59 & 0.29 & 1.29 \\
GPT3.5-turbo & 35.4{\tiny$\pm$5.6} & 0.51 & 0.45 & 2.11 & 40.3{\tiny$\pm$8.9} & 0.65 & 0.33 & 1.49 \\ 
GPT-4 & \textbf{43.1}{\tiny$\pm$5.9} & 0.61 & 0.44 & 1.89 & \textbf{41.9}{\tiny$\pm$8.9} & \textbf{0.74} & 0.31 & 1.28 \\ \midrule
Delphi & --- & --- & --- & --- & 22.6{\tiny$\pm$11.3} & --- & --- & --- \\
\bottomrule
\end{tabular}}
\end{subtable}
\begin{subtable}[t]{\textwidth}
\label{tab:original_story_prompt}
\caption{Zero-shot with Social Simulacra and Prompt Optimization}
\label{tab:prompt_methods}
\setlength\tabcolsep{2.7pt}
        \centering
         \resizebox{0.85\textwidth}{!}{
\begin{tabular}{@{}r|cccc|cccc@{}}
\toprule
Models & \multicolumn{4}{c|}{Causal (GPT3.5-davinci-v3)} & \multicolumn{4}{c}{Moral (GPT3.5-davinci-v2)} \\ \midrule
Methods                         & Agg ($\uparrow$)              & AUC ($\uparrow$)  & MAE ($\downarrow$) & CE ($\downarrow$)   & Agg ($\uparrow$)              & AUC ($\uparrow$)  & MAE ($\downarrow$) & CE ($\downarrow$) \\ \midrule 
Persona (average) & \cb{39.7}{\tiny $\pm$1.1} & 0.60 & 0.31 & 0.73 & \cb{28.9}{\tiny $\pm$1.6} & 0.55 & 0.33 & 0.86\\
Persona (best) & \br{42.7}{\tiny $\pm$5.6} & 0.60 & 0.31 & 0.71 & \br{37.1}{\tiny $\pm$8.5} & 0.58 & 0.27 & 0.72\\
Persona (worst) & \cb{37.2}{\tiny $\pm$5.4} & 0.62 & 0.29 & 0.61 & \cb{18.5}{\tiny $\pm$6.5} & 0.58 & 0.46 & 0.71 \\ \midrule
Auto Prompt Engineer & \cb{40.6}{\tiny $\pm$5.8} & 0.64 & 0.41 & 1.55 & \br{40.4}{\tiny $\pm$8.8} & 0.46 & 0.28 & 0.63\\
\bottomrule
\end{tabular}}
\end{subtable}
\end{table}

\paragraph{Zero-shot Alignment}

We report the result in \autoref{tab:original_story_exp}. We notice that there is no model that perfectly aligns with human judgments on these metrics. We first find that larger models (GPT-3, GPT-3.5, GPT-4, and Claude) generally align with humans better than smaller models (GPT-2, RoBERTa, etc.). This is most apparent when one compares the performance of GPT3-curie-v1 (6.7B) to GPT3-davinci-v1 (175B). Both models are trained using the same setup, but on both the \mo and the \ca, GPT3.5-davinci-v2 aligns better than GPT3-curie-v1.
While existing literature has shown improved performances of LLMs on general deterministic tasks~\citep{wei2022emergent}, we show here that LLM alignment also improves on ambiguous tasks.

Recently, techniques such as supervised fine-tuning on human instructions (GPT3.5-davinci-v2; \citealp{ouyang2022training}), using reinforcement learning to update LLM weights from human feedback (RLHF; GPT3.5-davinci-v3, GPT3.5-turbo, Anthropic-claude-v1; \citealp{stiennon2020learning,bai2022training}), and multi-modal joint training on both language and images (GPT-4; \citealp{openai2023gpt}) have improved LLMs on various tasks. Perhaps surprisingly, unlike the monotonic improvement we find by increased model size, these techniques impact these models' aggregate-level alignment differently.
Comparing the result between GPT3.5-davinci-v2 and GPT3.5-davinci-v3, RLHF seems to increase aggregate alignment on causal judgment but decrease alignment on moral judgment. Comparing the results between Anthropic-claude-v1 and GPT3.5-davinci-v3, it seems that even though both models are fine-tuned with RLHF, on aggregate, their alignment with human judgments differs. 
In the next section, we will design a method to examine how different training methods cause subtle shifts in the model's implicit tendency.

\paragraph{Social Simulacra}

\cite{park2022social} proposed a framework to elicit diverse social behaviors from LLMs by prompting them with different personas. For example, adding the persona, ``Jane Smith is a liberal activist.'' before the question. The human responses we collect on each story are diverse (see \autoref{fig:p_yes} for the distribution of P(Yes)). We shouldn't expect LLMs to generate a single response that matches the diversity of human perspectives. Therefore, we designed an experiment where we sample 5 personas from 5 simulacra defined by \cite{park2022social} (25 in total). We elicit $P$(``Yes''$|$Persona+Story+Prompt) and P(``No''$|$Persona+Story+Prompt) from the model. 

We report three ways of incorporating personas: Persona (average), where we average $P$ over all personas. Persona (best), where we find the persona that gave the most human-aligned responses. Persona (worst), where we use $P$ from the persona that gave the least human-aligned responses.
See how we designed this experiment in Appendix~\ref{sec:app:social_simulacra} and our sampled personas in \autoref{tab:persona}. We chose the GPT3.5-davinci-v2 and GPT-3.5-davinci-v3 for this experiment because their API endpoints explicitly provide the probability of Yes or No answers, and they align with humans best.

Personas elicit very different \emph{moral permissibility} responses -- the most and least-aligned persona have close to 18.6\% difference in discrete agreement. However, \emph{causal judgments} do not vary much based on different personas. We show the most and least aligned persona in Appendix~\ref{sec:app:social_simulacra}. 

\paragraph{Automatic Prompt Optimization}

We additionally explore whether we can use a black box method to automatically identify a prompt that is maximally aligned with collected human responses. \cite{zhou2022large} proposed Automatic Prompt Engineer (APE), a method that uses LLMs to generate prompts and then use the log-probability of a few training examples to select the best prompts. Note that APE does not currently have a workflow defined for chat-based models. Our experiment used five stories as training examples to select prompts. We describe how we searched the optimal prompt in Appendix~\ref{sec:app:ape}. We see that APE is still limited by the generative capability of LLMs -- we are unable to find a better prompt for the \ca for GPT3.5-davinci-v3. However, for the \mo, we found a prompt that substantially improved GPT3.5-davinci-v2's discrete agreement by 8.1\%. The discovered prompt is ``James is a consequentialist.'' On average, we found that prompts for the \mo with words like ``consequentialist'' or ``utilitarian'' lead to the highest alignment with humans. However, this observation is nuanced, and we provide an extended discussion in Appendix~\ref{sec:app:utilitarian}. For causality, good prompts are with the words ``a good student'', ``a good researcher'', or ``a very logical person''. We report all the generated prompts in Appendix~\ref{sec:app:ape}.

\subsection{Do LLMs have the same implicit tendencies as people on these stories?}

In the previous section, we focus on analyzing aggregate metrics such as agreement over all stories. Such analysis often provides no information beyond comparing highly complicated systems with a single number. 
Since each of our stories is a combination of factors with corresponding attributes (see \autoref{fig:story_example}), we can leverage conjoint analysis and compute the Average Marginal Component Effect (AMCE) for each factor attribute~\citep{hainmueller2014causal,awad2018moral}, where AMCE reveals the implicit tendency of the underlying system when a particular attribute is present.

\subsubsection{Method}

\paragraph{Average Marginal Component Effect}

We provide an overview of how AMCE is computed for each factor that reveals the implicit tendency of the system. For $N$ responses, $K$ stories, and $J$ factors. Each factor has 2 attributes/levels. For 3-level factors such as ``Time'', we only pick ``Early'' and ``Late'' attributes. Each response is $Y_{jnk} \in \{0, 1\}$, denoting $n$-th response, $j$-th factor, and $k$-th story. For LLMs, $Y_{jnk} = P($Yes$|$Story$)$.
We also define a factor profile matrix $T \in \{0, 1\}^{J \times K}$, where $T_{jk}$ denotes the attribute level of the $j$-th factor in $k$-the story.
Now, we can use a non-parametric difference-in-means estimator to compute AMCE:

\begin{equation}
\Delta(j) = \frac{ \sum_{j=1}^J \sum_{n=1}^N \sum_{k=1}^{K} \mathbbm{1}\{T_{kj} = 1\}Y_{jnk} 
}{ \sum_{j=1}^J \sum_{n=1}^N \sum_{k=1}^{K}  \mathbbm{1}\{T_{kj} = 1\}} - \frac{ \sum_{j=1}^J \sum_{n=1}^N \sum_{k=1}^{K}  \mathbbm{1}\{T_{kj} = 0\} Y_{jnk}}{ \sum_{j=1}^J \sum_{k=1}^{K} \sum_{n=1}^N  \mathbbm{1}\{T_{kj} = 0\}}  
\end{equation}

Note that if there is no preference between the two attributes of the same factor, $\Delta(j)=0$. Here is an example of how this estimator works. If we want to know if a system has a stronger tendency to find causing harm to be more morally permissible when they are ``Inevitable'' than ``Avoidable'' for the factor ``Counterfactual Evitability'', 
we can find all stories where ``Counterfactual Evitability'' has the attribute of ``Inevitable'', compute the average of all the binary responses $Y$ for all respondents, then subtract the average responses $Y$ for all stories with attribute ``Avoidable''. 
We additionally compute the 95\% confidence interval on ACME using Bootstrap and show these results in \autoref{fig:factor_distribution} and \autoref{fig:method_factor_distribution}. We only plot the average $\Delta(j)$ in \autoref{fig:spider_causal} and \autoref{fig:spider_moral}.
More details in Appendix~\ref{sec:app:amce}.

\paragraph{Zero-shot Implicit Tendency}

We use the standard radar chart to capture the multi-factor tendency of each LLM and human. We put the attributes of the same factor (e.g., Abnormal and Normal) on opposite sides -- since the leaning is over one attribute or the other: $\Delta(j) \in [-1, 1]$.
Each concentric circle is marked with a probability (e.g., 0.1), which represents the change in $P$(Yes) if the factor attribute had been present in the story. The implicit tendency is defined as the expected change in the probability of the system outputting ``Yes'' if this attribute had been present in the story. Intuitively, we can say a model has an implicit tendency for one attribute when it's more likely to judge an event to be the cause, or the impending harm to be more morally permissible, if this attribute is present.

\begin{figure}[b]
  \centering
  \includegraphics[width=0.9\linewidth]{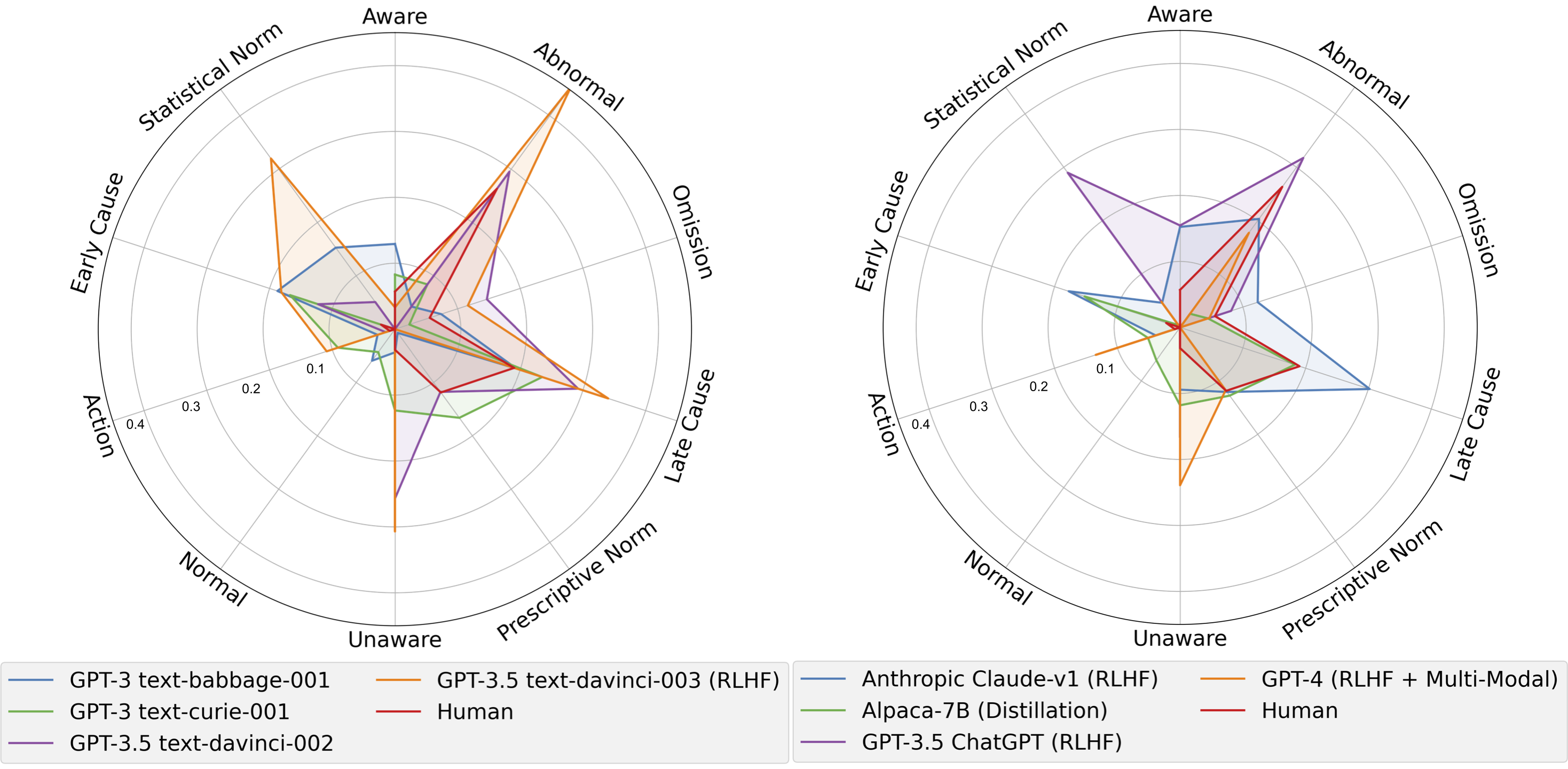}
  \caption{\textbf{Causal Implicit Tendencies}. Each concentric circle marks the implicit tendency -- the change in probability of responding ``Yes'' if the attribute had been present in the story. For example, if ``Abnormal" is present in the story, humans will have a 25\% higher probability to respond ``Yes'' on average than ``Normal'' being present. The left figure compares models of different sizes. The right figure compares models trained or finetuned with different methods.}
  \label{fig:spider_causal}
\end{figure}

\subsubsection{Results}

\paragraph{Causality: Sensitivity to Abnormality} %
People tend to cite abnormal events as the cause of the outcome. In our stories, the normality of events is clearly stated. However, a model still needs to first, identify the norm specified in the story and then, determine which event violated the norm. We find that text-davinci-003 exhibits a strong tendency for citing abnormal events as causal compared to any other models (and to humans). We also find that smaller models (text-babbage-001, text-curie-001, and Alpaca-7B) don't have a tendency for abnormal over normal events, which might indicate an inability to perform the required chained reasoning. 

\paragraph{Causality: Statistical or Prescriptive Norm}
Statistical norms refer to commonly and naturally occurring events, such as ``David usually comes home around 6 pm.'' or ``When you flip the switch, lights usually turn on.'' Prescriptive norms refer to human-made rules, such as ``You are not allowed to run in the hallway.'' or ``The speed limit is 60 mph.'' People tend to find events that violate prescriptive norms to be the cause of the outcome. Surprisingly, although GPT-4, Claude-v1, text-davinci-002, and smaller models, such as text-curie-001 and Alpaca-7B, have similar tendencies to humans, text-davinci-003 and its counterpart ChatGPT exhibit a strong tendency for statistical norms. This shows that LLM implicit tendencies can change for different training methods.

\begin{figure}[t]
  \centering
  \includegraphics[width=0.9\linewidth]{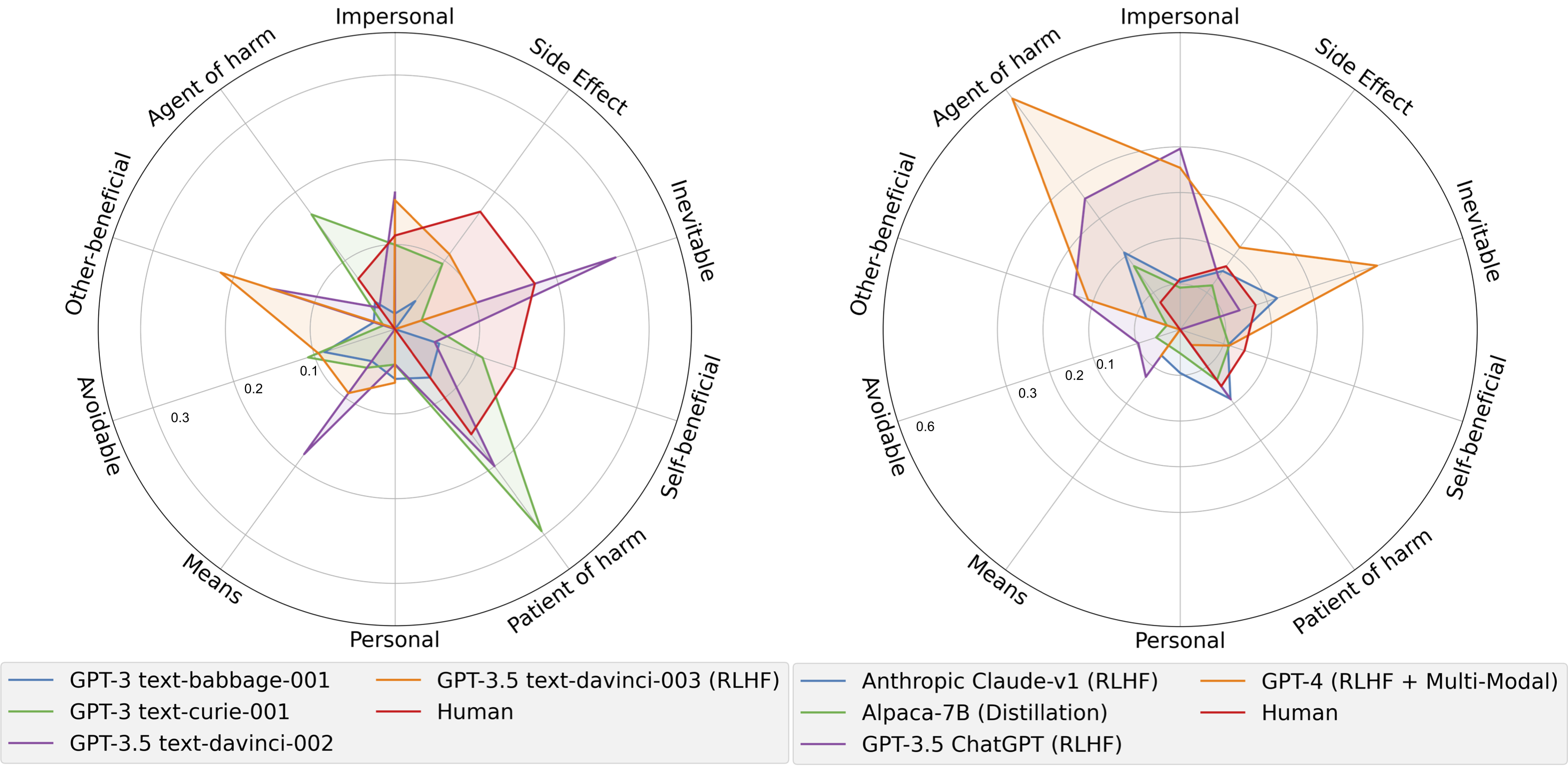}
  \caption{\textbf{Moral Implicit Tendencies}. Note that the Left and right figures have different scales.}
  \label{fig:spider_moral}
\end{figure}

\paragraph{Morality: Benefit Self or Others}
The stories in the \mo often involve scenarios where participants must make a choice: they are either safe from danger, and their decision only affects others, or they need to save their own lives along with other people. 
Humans are more likely to save themselves (self-beneficial). Interestingly, smaller models such as text-babbage-001 and text-curie-001 find actions more morally permissible when they are self-beneficial, while larger models do not. 
RLHF makes the models more other-beneficial. For example, while Claude-v1 seems to prefer more self-beneficial actions, text-davinci-003, ChatGPT, and GPT-4, all prefer to be other-beneficial. Human and model tendencies do not align on this factor, but since models are lifeless systems, we may want models to be more other-beneficial.
See \autoref{fig:factor_distribution} for a clearer comparison.

\paragraph{Morality: Inevitable or Avoidable Harm}
Some stories involve situations where harm would inevitably happen to a group of people regardless of the participant's choice. Harm directed towards inevitable consequences is often considered less bad. This requires the model to comprehend the information described in the story about the counterfactual scenario. We find that ChatGPT, Claude-v1, and GPT-4 find harmful actions toward inevitable consequences much more morally permissible than any other model.

\paragraph{Morality: Intervening on Agent or Patient of Harm}
A hijacked airplane full of passengers is about to hit a building full of people. The people in the hijacked airplane are referred to as agents of harm, and the people in the building are the patients of harm.
Since we compute the confidence interval on AMCE, although some models (such as GPT-4) seem to have a tendency to intervene on the agent of harm, the upper/lower bound of the confidence interval shows there might not be a clear tendency (see \autoref{fig:factor_distribution}).

\paragraph{Morality: Personal or Impersonal Force} This factor captures whether or not the outcome was brought about by using personal force (e.g., pushing a person). Generally, people are less likely to find actions morally permissible that involve personal force. We find that only two models, ChatGPT and GPT-4, align with human intuitions about personal force.

\paragraph{Morality: Means or Side Effect} Assessing an agent’s causal role is important for moral judgments. Generally, people find actions more morally permissible when they lead to harm as side effects rather than when the harm is necessary to bring out the outcome. This can be illustrated as a diagram in \autoref{fig:means_sideeffect}. We find that no model exhibits a clear preference for harm as a side effect.

We additionally saw the following larger trends across models:
\begin{enumerate}[itemsep=0pt, wide=0pt, leftmargin=*, after=\strut,label={(\arabic*):}]
    \item \textbf{Non-monotonicity}: The alignment to human biases does not necessarily increase with model size. We speculate that alignment is an area where the inverse scaling law applies~\citep{mckenzie2023inverse}.
    \item \textbf{Heterogeneity}: Interestingly, but perhaps not surprisingly, models that used the same training method and fine-tuned for human preferences do not have the same implicit tendencies -- we highlight the difference between Claude-v1 and GPT3.5-turbo.
    \item \textbf{Self-Tendency vs. Other-Preference}: Humans are ego-centric and often make self-beneficial decisions. However, when asked to judge the behaviors of others, we prefer others to be altruistic. For example, in scenarios involving self-driving cars, \cite{kallioinen2019moral} reported a stark contrast between what we tend to do versus what we want other people to do. This difference will make models trained on human preferences through RLHF different from how humans would tend to act in a given situation. As it is more prevalent to use LLMs as proxy humans for data labeling or experiments~\citep{dillion2023can,gilardi2023chatgpt}, it is increasingly important to understand and measure these implicit differences.
\end{enumerate}

\subsubsection{Other Analyses}

\paragraph{Same Model with Different Prompting Methods} We also investigate how different prompting methods shift a model's implicit tendencies. Prompting methods shifted the model's implicit tendency significantly for some factors (e.g., abnormal sensitivity, benefit self or others) but not much for others (e.g., inevitable or avoidable harm). We report this in \autoref{fig:method_factor_distribution}.

\paragraph{Predicting Factor-Attribute as Few-shot Learning} We asked LLMs to predict which factor attribute each story contains as a few-shot natural language understanding task. Across all prediction tasks, text-davinci-v2 achieved 85.5\% accuracy on causal factors and 70.9\% on moral factors. We report more details in Appendix~\ref{sec:app:factor_infer}. This shows that even though LLMs and humans do not align, they are very capable of following examples and instructions to annotate stories with factors that are relevant to human judgments.

\paragraph{Hallucination}

\begin{wraptable}{r}{5cm}
\caption{\textbf{Model Hallucination}: number of stories where the model hallucinated.}
\begin{tabular}{lcc}
\toprule
Model          & Causal & Moral \\
\midrule
text-curie-001 & 8/10   & 3/10  \\
Claude-v1      & 2/10   & 2/10  \\
GPT3.5-turbo  & 2/10   & 0/10  \\
GPT-4          & 0/10   & 0/10  \\
\bottomrule
\end{tabular}
\label{tab:hallu}
\end{wraptable}

We conduct a small-scale analysis to investigate whether the tendency difference is due to actual tendency difference or model hallucinations.
We annotated 80 examples by sampling 10 examples where 4 models made mistakes across 2 tasks. We prompt the model to explain why they made the choice for the original story. We use the explanation to check if the model hallucinates and has an accurate grasp of the facts represented in the story.
A model hallucinates when they re-state the core story in a way that’s inconsistent with the facts provided in the story. For example, if an action is not performed by character A, and the model thinks it is performed by character A, we count this as a hallucination. In \autoref{tab:hallu}, we can clearly see that when a model makes a mistake, smaller models tend to hallucinate more.

We read the model explanations on examples where no hallucination is found, and we conclude that larger models indeed have a different tendency compared to humans in these stories. On moral stories, Claude-v1, GPT3.5-turbo, and GPT-4 all seem to be bound by pre-entered moral principles and choose the same action regardless of the story circumstances (potentially due to the influence of AI constitution; \citealp{bai2022constitutional}), while humans tend to consider a few more factors (i.e., nuances in the story). For example, when faced with intricate situations, language models tend to take a very passive stance. This is true even if the action could save a greater number of people by putting a smaller number of people in danger. However, when self-sacrifice is in question, the language model is inclined to metaphorically ``sacrifice'' itself. Our observation is based on a limited set of examples. We leave more extended evaluation to future work.

%% file: sections/discussion.tex
\vspace{-3mm}
\section{Conclusion}
\vspace{-3mm}

We summarized the main findings of 24 cognitive science papers around human intuitions on causal and moral judgments.
We collected a dataset of causal and moral judgment tasks and used the scientific findings to enrich the stories with structural annotations. 
Using conjoint analysis, we compute the average marginal component effect (AMCE) for people's and the models' implicit tendency on the factors important to making causal and moral judgments. We find subtle differences in what humans and models prefer. Our work illuminates the importance of using a richly annotated dataset to understand a model's systematic tendencies, an important step to understanding where models and humans align or misalign.

\vspace{-3mm}
\section*{Acknowledgments}
\vspace{-3mm}

Research reported in this paper was supported in part by a Hoffman-Yee grant and a HAI seed grant. We would like to thank Paul Henne for his assistance in curating high-impact causal reasoning literature. We thank Lucy Li for writing the ethical consideration section for this paper. We would also like to thank Jon Gauthier, Nicholas Tomlin, Matthew Jorke, Xuechen Li, Sameer Jha, Rose E Wang, and Dora Demszky for general discussions. An earlier version of the dataset was submitted to BigBench. We would like to thank the reviewers Marcelo Menegali, Adrià Garriga-Alonso, and Denis Emelin for their comments and feedback.
We thank May Khoo for data annotations and facilitation of labeling. We thank Simon Lee Huang for his work to run experiments with Automated Prompt Engineer and Social Simulacra.

%% file: sections/appendix.tex
\section{Appendix}

\subsection*{Ethical considerations}
\vspace{-3mm}

It is imperative that we assess implicit intuitions underlying commonsense reasoning abilities in LLMs capable of open language generation. This is especially important for cases related to morality, considering that algorithmic neutrality is not possible: systems reflect the data they ingest and the people who build them \citep{green2021data}. In addition, even if a model is not explicitly given the responsibility to make moral judgments, these judgments can appear across many forms of freely generated text. Some may argue that equipping models with moral reasoning skills can allow them to behave better in complex environments, but an AI capable of reasoning is not necessarily bound to be ethically well-aligned, as many humans themselves have demonstrated \citep{cave2019ethics,talat2021word}. 
Using carefully curated data, we advocate for the inspection of \textit{how} choices are made and whether factors used in human cognitive processes are or can be incorporated by models. 
In the case of our moral permissibility task, we would like it to be a dataset for investigating these underlying factors rather than a flat benchmark to beat. 
Whether alignment between LLMs and human participants is a desired goal in this context is also unclear. For example, there are well-documented biases that affect people's moral intuitions, and we wouldn't want to replicate these biases in LLMs \citep{eberhardt2020biased}.
However, we argue that despite these challenges and difficulties, our research, which focuses on utilizing the insights and frameworks proposed by cognitive science and philosophy to analyze and scaffold LLMs to generate more human-aligned judgments, is a small step forward in a promising direction.

\subsection{Factors in Causal Selection Task}
\label{sec:app:causal_factors}

\paragraph{Causal Structure}
In a situation where multiple events are present in a story, and the outcome is caused by one or more of these events, it's important to understand the causal structure of the situation: Is one event sufficient to bring about the outcome (disjunctive causes) or are all of the considered events required (conjunctive causes)?~\citep{wells1989mental,mandel2003judgment,sloman2015causality}. 

\paragraph{Event Normality}
People have a general tendency to cite abnormal events (rather than normal events) as causes~\citep{knobe2008causal,hitchcock2009cause,alicke2011causation,o2021degrading,morris2019quantitative}. This effect is present for both children and adults~\citep{samland2016role}. Norms can be broadly classified into ``statistical norm'' (norms about what tends to be) or ``prescriptive norm'' (norms about what should be)~\citep{icard2017normality,n1995mental,kominsky2015causal}. Event normality applies to inanimate objects as well. For example, a malfunctioning machine is judged by more people to be the cause than a machine that functions as it should (often called the ``norm of proper functioning'')~\citep{sytsma2021crossed}.

\paragraph{Agent knowledge}
The epistemic state of the agent also plays an important role in how people judge whether the agent is the cause of the outcome. For example, an agent who is aware of the potential consequences of their action and knowingly performs an action that leads to a foreseeable bad outcome is judged more harshly than an agent who lacks the relevant knowledge. This is often characterized as knowledge versus ignorance~\citep{samland2016role,kominsky2019immoral}. 

\begin{table}
\caption{Factors that influence causal selection judgments (left) and moral permissibility judgments (right).  We provide a list of simplified definitions for each tag in the story. We also added references to papers that primarily investigate these factors. We refer to higher-level concepts (e.g., ``Causal Role'') as factors and the possible values for each factor (e.g., ``Means'' or ``Side Effect'') as tags.}
\vspace{0.1cm}
\label{tab:causal_tag_defs}
\begin{minipage}[t]{0.53\textwidth}
\strut\vspace*{-\baselineskip}\newline
{\fontsize{8.5}{9.5}\selectfont
\resizebox{\columnwidth}{!}{
\begin{tabular}{r  p{0.6\columnwidth}}
\toprule
\multicolumn{2}{c}{\textbf{\textsc{Causal selection}}} \\
\midrule
Factors & Definitions\\
\midrule
\begin{tabular}[t]{@{}r@{}} \textbf{Causal} \\ \textbf{Structure}\end{tabular} &   \citep{wells1989mental,mandel2003judgment,sloman2015causality}\\
\midrule
Conjunctive & All events must happen in order for the outcome to occur. Each event is a  necessary cause for the outcome. \\
Disjunctive & Any event will cause the outcome to occur. Each event is a sufficient cause for the outcome.\\
\midrule
\begin{tabular}[t]{@{}r@{}} \textbf{Agent} \\ \textbf{Awareness}\end{tabular} &    \citep{samland2016role,kominsky2019immoral}   \\
\midrule
Aware                                         & Agent is aware that their action will break the norm/rule or they know their action is ``abnormal''.                         \\
Unaware                                       & Agent is unaware or ignorant that their action will break the norm/rule, or they don’t know their action is ``abnormal''.   \\
\midrule
\textbf{Norm Type}   & \citep{icard2017normality,n1995mental,kominsky2015causal, sytsma2021crossed}\\
\midrule
Prescriptive Norm                             & A norm about what is supposed to happen. \\
Statistical Norm                              & A norm about what tends to happen.    \\ 
\midrule
\begin{tabular}[t]{@{}r@{}} \textbf{Event} \\ \textbf{Normality}\end{tabular}  &  \citep{knobe2008causal,hitchcock2009cause,alicke2011causation, o2021degrading,morris2019quantitative,samland2016role}\\
\midrule
Normal Event                                  & The event that led to the outcome is considered ``normal''.\\
Abnormal Event                                & The event that led to the outcome is considered ``abnormal/unexpected''.       \\
\midrule
\begin{tabular}[t]{@{}r@{}} \textbf{Action} \\ or \\ \textbf{Omission}\end{tabular}  &
\citep{ritov1992status,baron2004omission,henne2017cause,henne2019counterfactual,clarke2015causation,descioli2011omission,gerstenberg2021omission}                           \\
\midrule
Action as Cause                               & Agent performed an action that led to the outcome. \\
Omission as Cause                             & Agent did not perform the action, and the omission led to the outcome. \\ 
\midrule
\textbf{Time}                  &  \citep{reuter2014good,henne2021counterfactual}                                                                                                                                                                                                                    \\
\midrule
Early                                         &  The event happened early.\\
Late                                          & The event happened late.\\
Same Time                                     & Multiple events happened at the same time. \\
\bottomrule
\end{tabular}
}}
\end{minipage}
\begin{minipage}[t]{0.47\textwidth}
\strut\vspace*{-\baselineskip}\newline
{\fontsize{10}{11}\selectfont
\resizebox{\columnwidth}{!}{
\begin{tabular}{r  p{0.6\textwidth}}
\toprule
\multicolumn{2}{c}{\textbf{\textsc{Moral permissibility}}} \\
\midrule
Factors                                       & Definitions  \\ 
\midrule
\textbf{Causal Role}        &  \citep{hauser2006moral,christensen2014moral,wiegmann2012order} \\
\midrule
Means                                       & The harm is instrumental/necessary to produce the outcome.  \\
Side Effect                                         & The harm happens as a side-effect of the agent's action, unnecessary to produce the outcome. \\
\midrule
\textbf{Personal Force}        &                                                                                        \citep{christensen2014moral,wiegmann2012order} \\
\midrule
Personal                                           & Agent is directly involved in the production of the harm. \\ %
Impersonal                                         & Agent is only indirectly involved in the process that results in the harm (e.g., using a device). \\ %
\midrule

\begin{tabular}[t]{@{}r@{}} \textbf{Counterfactual} \\ \textbf{Evitability}\end{tabular}   &     \citep{moore2008shalt,huebner2011source,christensen2014moral,wiegmann2012order} \\
\midrule
Avoidable                                          & The harm would not have occurred if the agent hadn't acted.  \\
Inevitable                                         & The harm would have occurred even if the agent hadn't acted.  \\
\midrule
\textbf{Beneficence}          &         \citep{bloomfield2007morality,christensen2014moral,wiegmann2012order} \\
\midrule
Self-Beneficial                                    & The agent themself benefits from their action.  \\ 
Other-Beneficial                                   & Only other people benefit from the agent's action.  \\
\midrule
\begin{tabular}[t]{@{}r@{}} \textbf{Locus of} \\ \textbf{Intervention}\end{tabular}  &     \citep{waldmann2007throwing} \\
\midrule
Instrument of Harm                                 & The intervention is directed at the instrument of harm (e.g., the runaway train, the hijacked airplane). \\
Patient of Harm                                    & The intervention is directed at the patient of harm (e.g., the workers on the train track). \\ 
\bottomrule
\end{tabular}
}}
\end{minipage}
\end{table}

\paragraph{Action/omission} 
People tend to select an action as the cause rather than inaction. This phenomenon is often called the ``omission bias''~\citep{ritov1992status,baron2004omission}. In a scenario where one person acts, and another one doesn't, the person who acted tends to be cited as the cause of the outcome \citep{henne2017cause,henne2019counterfactual,clarke2015causation,descioli2011omission,gerstenberg2021omission}.

\paragraph{Temporal effect}
Causal selections are also affected by the order in which the events occur \citep{gerstenberg2012when}. When several events unfolding over time lead to an outcome, people have a general tendency to select later events rather than earlier events as the actual cause of the outcome~\citep{reuter2014good}. However, it also depends on how the events are causally related to one another \citep{hilton2010selecting,spellman1997crediting}. When earlier events determine the course of action, these events tend to be selected ~\citep{henne2021counterfactual}. 

\subsection{Factors in Moral Permissibility Task} 
\label{sec:app:moral_factors}

\begin{figure}[b]
  \centering
  \includegraphics[width=0.6\linewidth]{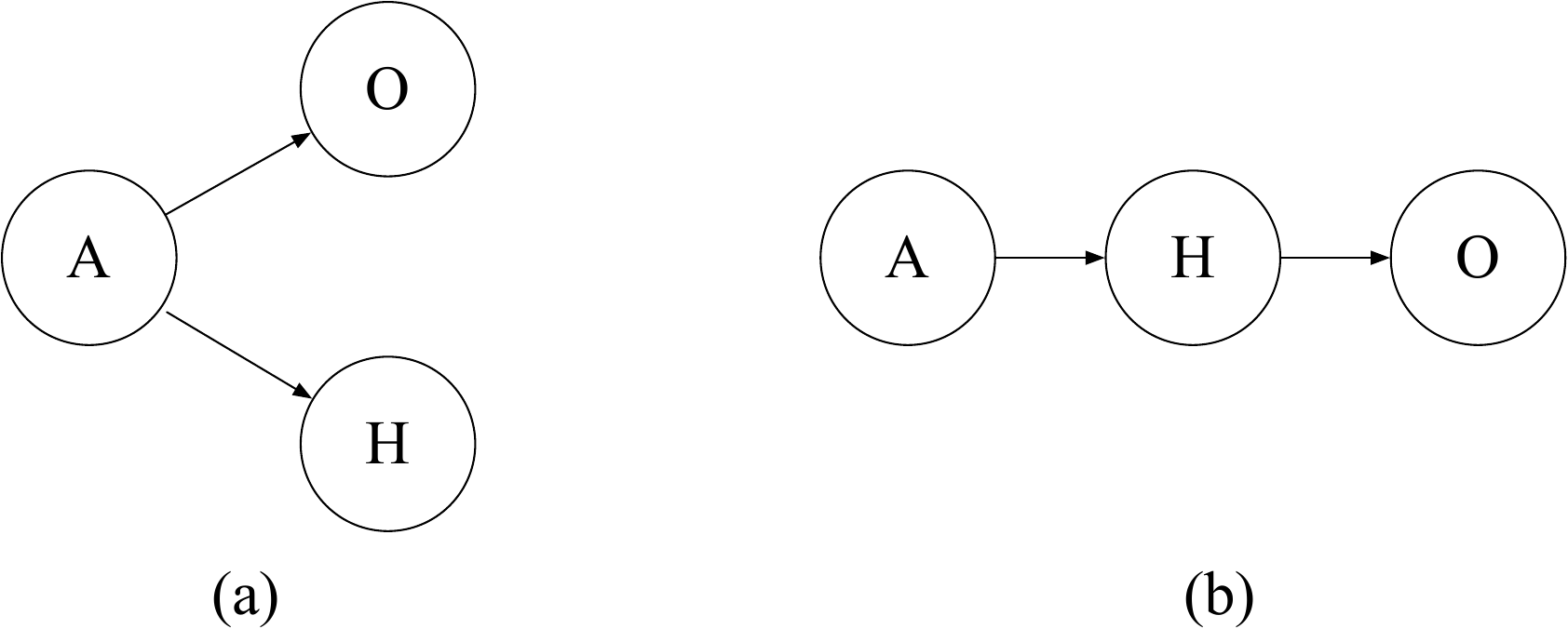}
  \caption{\textbf{Causal Role} We show the causal diagram of the difference between means and side effects. A represents action conducted by the agent. H represents the harm. O represents the outcome. (a) is the diagram that represents harm as a side effect, and (b) represents harm as a means.}
  \label{fig:means_sideeffect}
\end{figure}

\paragraph{Causal Role} Assessing an agent's causal role is important for moral judgments. Actions are more likely to be seen as permissible when they didn't play an important causal role in how the negative outcome came about. One important distinction is between an event as a means versus a side effect of the outcome (illustrated in \autoref{fig:means_sideeffect}). In \autoref{fig:means_sideeffect}a, the action (A) leads to a desired outcome (O) but also has the side effect of harming someone (H). In \autoref{fig:means_sideeffect}b, the harm is a means to bringing about the outcome. Generally, people find actions more morally permissible when they lead to harm as side effects rather than when the harm is a means for bringing about the outcome. 

\paragraph{Personal Force}

This factor captures whether or not the outcome was brought about by the use of personal force (e.g., pushing a person). Generally, people are less likely to find actions morally permissible that involve personal force. 

This is also a factor that connects causal judgment and moral judgment -- in causal judgment, a normative framework that describes whether a cause will be selected by human tendency is through the judgment and observation of force transfer~\citep{wolff2007representing}. Moral philosophy that studied the mass murders during the Second World War has suggested that killing via a mechanism (e.g., ``press a button'') is much easier than actively involved in the killing acts~\citep{eatherly2015burning}.

\paragraph{Counterfactual Evitability}

Could the harm have been avoided? In the story where the teenager was going to trigger the bomb, which would kill him and the others in the building, in a counterfactual world, had the agent not killed the teenager, they would have died from the explosion they caused anyway. Harm directed towards inevitable consequences is often considered less evil~\citep{hauser2006moral}.

\paragraph{Beneficence}

People are more likely to take actions that benefit themselves~\citep{bloomfield2007morality}. In the stories, participants were not asked to choose themselves over others; instead, the framing often is they are either safe from danger, and their decision only affects others, or they need to save their own lives along with other people.

\paragraph{Locus of Intervention}

People are often asked to choose to intervene on the instrument of harm or recipient of harm.
For example, a hijacked airplane with passengers can crash into a building full of people. If we use a missile to shoot down the hijacked airplane, killing all the passengers onboard, then we would be intervening on the instrument of harm (the plane). If we choose to misdirect the hijacked airplane's navigation system to hit another smaller building also filled with fewer people, then we would be intervening on the patient of harm. Presented with this type of choice, human tendency is usually to intervene on the instrument of harm~\citep{waldmann2007throwing}.

\subsection{Original Story Experiment Details}
\label{sec:app:r1_exp_details}

We conduct a 3-class comparison, a binarized response of ``Yes'' or ``No'', and an additional class of ``ambiguous'' when the agreement between human or model output is 50\% ± 10\%.
We test alignment using a prompt-based zero-shot task setup.
For each model and scenario, we use the normalized probability and compare $P(\text{``Yes''}|\text{Story + Prompt})$ and $P(\text{``No''}|\text{Story + Prompt})$. 
We apply temperature-based sampling to generate samples of ``Yes'' or ``No'' answers per story for each experimental run. 
For models that are trained with a masked language model objective, we append a mask token after the prompt question. For a generative language model like GPT2, we directly ask it to produce the next word. We also included two models for which we didn't have direct access -- GPT-3~\citep{brown2020language} and Delphi~\citep{jiang2021delphi}. We used their APIs to evaluate alignment. 

For Chat-based APIs (Alpaca-7B, Anthropic-claude-v1, GPT3.5-turbo, GPT-4), since the probability of each token is not provided, we first experimented with sampling multiple times from a model with a high temperature. We sampled 10 responses for a few stories to see if we have any variation in generated responses -- to our surprise, we always get the same response. After this initial exploration, we only generate one response per story and set $P = 1$ for the chosen response. In our prompt, we tell the chat-based models that they can output ``not sure'' if they are not certain. When ``not sure'' is found, we assign $P = 0.5$ for either choice.

\subsection{Social Simulacra Experiment Details}
\label{sec:app:social_simulacra}

There are 50 simulacra (community) in total provided by \cite{park2022social}; each simulacra has 200 persona prompts. The persona in each simulacrum roughly shares a common theme. Some of such themes include ``A community for discussing Yu-Gi-Oh! Master Duel!'' or ``A community for discussing Disney's movie, Encanto.''

Most of the simulacra themes are highly irrelevant to our task domain and evaluating all simulacra personas are cost prohibitive. Therefore, we choose 5 simulacra that are most relevant to us -- they contain personas that usually describe ideological or political perspectives. We then randomly choose 5 personas out of 200 personas in each simulacrum. We show our selection in \autoref{tab:persona}. You can access the complete list of 200 personas, such as personas listed in LuxuryLifeHabits here\footnote{\url{https://social-simulacra.herokuapp.com/LuxuryLifeHabits\#simulacra}}.

For both models, the persona that most closely aligned with our collected responses for \ca is ``Emily White is a Republican who got fed up with the party'', and the least aligned persona is ``Angela Campbell is a woman who likes to party until the sun comes up''.
For \mo, the most aligned persona for both models is ``Allen Lee is a really cool guy'', and the least aligned persona is ``Paul Brown is an anti-vaccine activist''.

\begin{table}[h]
\small
\centering
\caption{The full zero-shot performance of two GPT3.5 models with Social Simulacra and Prompt Optimization methods.}
\label{tab:full_social_sim}
\begin{subtable}[t]{\textwidth}
\caption{Zero-shot with Social Simulacra and Prompt Optimization (text-davinci-002)}
\label{tab:prompt_methods_davinci002}
\setlength\tabcolsep{2.7pt}
         \resizebox{\textwidth}{!}{
\begin{tabular}{@{}r|cccc|cccc@{}}
\toprule
\multicolumn{1}{l}{}           & \multicolumn{4}{c}{\textbf{Causal Judgment}} & \multicolumn{4}{c}{\textbf{Moral Permissibility}} \\ \midrule
Methods                         & Agg ($\uparrow$)              & AUC ($\uparrow$)  & MAE ($\downarrow$) & CE ($\downarrow$)   & Agg ($\uparrow$)              & AUC ($\uparrow$)  & MAE ($\downarrow$) & CE ($\downarrow$) \\ \midrule 
GPT3.5-davinci-v2             & 37.8{\tiny$\pm$5.6} & 0.61  & 0.31 & 0.72 & {32.3}{\tiny$\pm$8.1} & 0.67 & 0.30 & 0.74 \\ \midrule \midrule
Persona (average) & 39.7{\tiny $\pm$1.1} & 0.60 & 0.31 & 0.73 & 28.9{\tiny $\pm$1.6} & 0.55 & 0.33 & 0.86 \\
Persona (best) & 42.0{\tiny $\pm$5.9} & 0.60 & 0.30 & 0.71 & 37.1{\tiny $\pm$8.5} & 0.58 & 0.27 & 0.72 \\
Persona (worst) & 36.8{\tiny $\pm$5.9} & 0.62 & 0.29 & 0.61 & 18.5{\tiny $\pm$6.9} & 0.58 & 0.46 & 0.71 \\ \midrule
Auto Prompt Engineer & 38.5{\tiny $\pm$5.8} & 0.60 & 0.36 & 0.89 & 40.4{\tiny $\pm$8.8} & 0.46 & 0.28 & 0.63\\
\bottomrule
\end{tabular}}
\end{subtable}

\begin{subtable}[t]{\textwidth}
\caption{Zero-shot with Social Simulacra and Prompt Optimization (text-davinci-003)}
\label{tab:prompt_methods_davinci003}
\setlength\tabcolsep{2.7pt}
         \resizebox{\textwidth}{!}{
\begin{tabular}{@{}r|cccc|cccc@{}}
\toprule
\multicolumn{1}{l}{}           & \multicolumn{4}{c}{\textbf{Causal Judgment}} & \multicolumn{4}{c}{\textbf{Moral Permissibility}} \\ \midrule
Methods                         & Agg ($\uparrow$)              & AUC ($\uparrow$)  & MAE ($\downarrow$) & CE ($\downarrow$)   & Agg ($\uparrow$)              & AUC ($\uparrow$)  & MAE ($\downarrow$) & CE ($\downarrow$) \\ \midrule 
GPT3.5-davinci-v3 & 41.3{\tiny$\pm$5.6} & 0.67 &  0.41 & 1.39 & 20.2{\tiny$\pm$7.3} & 0.61 & 0.51 & 2.61\\ \midrule \midrule

Persona (average) & 39.6{\tiny $\pm$1.1} & 0.60 & 0.31 & 1.56 & 28.8{\tiny $\pm$1.6} & 0.55 & 0.33 & 0.86 \\
Persona (best) & 42.7{\tiny $\pm$5.6} & 0.60 & 0.31 & 0.71 & 37.1{\tiny $\pm$8.5} & 0.57 & 0.28 & 0.72 \\
Persona (worst) & 37.2{\tiny $\pm$5.4} & 0.62 & 0.29 & 0.61 & 18.5{\tiny $\pm$6.5} & 0.58 & 0.46 & 0.71 \\ \midrule
Auto Prompt Engineer & 40.6{\tiny $\pm$5.8} & 0.64 & 0.41 & 1.55 & 33.3{\tiny $\pm$8.3} & 0.50 & 0.47 & 1.62 \\
\bottomrule
\end{tabular}}
\end{subtable}

\end{table}

\begin{table*}
\centering
\small
\caption{List of sampled persona from the Social Simulacra Dataset}
\label{tab:persona}
\begin{tabular}{c|l}
\toprule
\textbf{Category} & \textbf{Persona Prompt} \\
\midrule
\multirow{5}{*}{insaneProtestors} & Amy Hsuis a lawyer with a former history working with social movements \\
 & Jared Cox is a firefighter who is married and has 3 children \\
 & Ryan Andrews is a photographer who likes to go on adventures \\
 & Tom Smith is a teacher who is preparing the next generation \\
 & Jane Smith is a liberal activist \\
\midrule
\multirow{5}{*}{luxuryLifeHabits} & Jake Williams is a person who has been abused \\
 & Polly Wangerin is a feminist and a democratic socialist \\
 & Jordan Sanders is a woman who supports Medicare for All \\
 & Emily White is a Republican who got fed up with the party \\
 & Nadia Rasheedis a scientist who is really good at science \\
\midrule
\multirow{5}{*}{newDealAmerica} & Nia Ellis is a Bernie Sanders supporter; a woman of color \\
 & Kayla James is a person who studied biology \\
 & Sarah Lee is a Bernie Sanders supporter who loves dogs \\
 & Jon Blair is the Director of the Office of LGBT Affairs \\
 & Sam Thompson is a retired union worker and Hillary Clinton supporter \\
\midrule
\multirow{5}{*}{nonPoliticalTwitter} & Angela Campbell is a woman who likes to party until the sun comes up \\
 & Liz Akins is a mom who wants to know how the plant will help her son \\
 & Tristan Wright is a person who likes to make new friends \\
 & Anthony Woods is a man who likes to meditate \\
 & Allen Lee is a really cool guy \\
\midrule
\multirow{5}{*}{politicsPeopleTwitter} & Suzy Han is a doctor trying to give mental health support to the alt-right \\
 & Dennis Johnson is a local minister fighting against misinformation \\
 & Paul Brown is an anti-vaccine activist \\
 & Ann Watson is a law student that has been trying to spread political awareness \\
 & Sam Thompson is a student trying to expose right-wing lies \\
\bottomrule
\end{tabular}
\end{table*}

\subsection{Automatic Prompt Engineer Experiment Details}
\label{sec:app:ape}

We use the official implementation of Automatic Prompt Engineer\footnote{\url{https://github.com/keirp/automatic_prompt_engineer}}. We asked LLMs to generate 200 prompts for GPT3.5-davinci-v2 and GPT3.5-davinci-v3.

Automatic Prompt Engineer provides us with a few default templates for engineering prompts. These prompts are fine if the prompt you’re trying to engineer can be described as an imperative statement, like ``produce X'' or ``generate Y.''
However, in our experiment, these default templates do not lead to any good prompts. Inspired by the Social Simulacra experiment results, we switched to a more persona-like template.

\begin{displayquote}
``James gave these answers to the following questions:

\textsc{[EXAMPLES]}

James is a \textsc{[PROMPT]}''
\end{displayquote}

We chose ``James'' because it is a common first name. We are aware that LLMs have implicit biases towards gendered names, and we suspect the generated prompt will be different if the name changes. We leave a systematic exploration of bias in automatic prompt generation for future work. 

In this prompt, \textsc{[Examples]} is replaced with input-output pairs from the dataset in the format of the default demos template, which is: \textsc{[Story] Output: [Answer]}.
After the prompt has been generated, APE evaluates the quality of the prompt by the following template:

\begin{displayquote}
``James is a \textsc{[PROMPT]}.

James is given this prompt: "\textsc{[STORY]}"

James is asked to please answer Yes or No.

James answers: \textsc{[Answer]}''
\end{displayquote}

We randomly sampled 5 causal and moral stories to use as training examples and excluded them from the experiment result. These stories are indices 1, 8, 23, 29, 112 for \ca and indices 0, 7, 28, 49, 61 for \mo. For evaluation and numbers reported in \autoref{tab:prompt_methods}, we select the best prompt for each model and use it to generate responses for all our stories.

We show the generated prompt and their average log-probability (the higher, the better):

\begin{verbatim}
Causal
text-davinci-002
----------------
-0.97:  very honest person, so his answers are all correct.
-1.01:  good student, and he always gives correct answers.
-1.02:  good student, and he always does his homework on time. 
        However, he is not very good at tests. 
        When he took his last math test, he got a C.
-1.08:  very good student, and he always gives correct answers.
-1.10:  good researcher, and he is careful to make sure that 
        his answers are correct.
-1.11:  very logical person, and he has thought about 
        these questions very carefully. 
        He is certain that his answers are correct.
-1.11:  very good student, and he always gets full marks on his tests.
-1.12:  good researcher, and he is very methodical. He is very careful
        to make sure that all of his answers are consistent with the 
        information given in the question.
-1.12:  very logical person, and he seems to be very sure 
        of his answers. I would trust his answers.
-1.13:  very literal person. He always answers questions 
        with a simple yes or no, regardless of whether 
        that is the most accurate answer.
\end{verbatim}

\begin{verbatim}
Causal
text-davinci-003
----------------
-0.87:  trustworthy source.
-0.98:  very smart person!
-1.25:  reliable answerer.
-1.27:  good listener!
-1.29:  responsible answer-giver.
-1.34:  very smart student!
-1.49:  reliable source of answers.
-1.51:  reliable answer giver.
-1.61:  reliable source of answers!
-1.63:  reliable source of accurate answers.
\end{verbatim}

\begin{verbatim}
Moral
text-davinci-002
----------------
-1.11:  consequentialist.
-1.15:  consequentialist, so he would always choose the option 
        that leads to the best outcome, even if it means sacrificing 
        one life to save many.
-1.17:  consequentialist, meaning he believes that the morally right 
        thing to do is the one that will produce the best consequences. 
        In each of the scenarios above, James believes that the best 
        thing to do is the one that will save the most lives.
-1.21:  utilitarian. He believes that the morally right course of action 
        is the one that results in the greatest good for the greatest 
        number of people. In the first scenario, releasing one 
        person from the helicopter will result in the greatest 
        good for the greatest number of people
-1.23:  utilitarian, so he would always choose the option that will result 
        in the greatest good for the greatest number of people. In the first 
        scenario, he would throw one person off the helicopter to save the 
        other eight. In the second scenario, he would drive
\end{verbatim}

\begin{verbatim}
Moral
text-davinci-003
----------------
-0.00:  utilitarian and believes in maximizing overall happiness. Therefore, 
        his answer to all of the questions is yes, as sacrificing the lives of 
        one or a few people for the sake of saving more lives is the action that 
        will bring the most happiness overall.
-0.00:  utilitarian, so he believes that the greatest good for the greatest 
        number should be pursued. Therefore, his answer to all of the questions 
        is yes, because in each case sacrificing a few people will save the 
        lives of many more people.
-0.01:  utilitarian, believing that decisions should be made to produce the 
        greatest good for the greatest number of people. His answers of "yes" in 
        all of these scenarios reflects his belief that it is best to sacrifice 
        a few individuals in order to save the many.
-0.01:  utilitarian and believes that the most ethical action is the one that
        results in the greatest good for the greatest number of people. In each 
        of these scenarios, sacrificing one person for the sake of saving more 
        lives is the best option according to utilitarianism, so
-0.01:  person who believes that the ends justify the means, as his answers 
        demonstrate. He believes that sacrificing the lives of a few to save the
        lives of many is justified in certain circumstances.
\end{verbatim}

\subsection{Additional Discussion on the Utilitarian Principle in Moral Judgments}
\label{sec:app:utilitarian}

There is a difference between what prompt aligns LLMs best with humans and the actual responses produced for each story. The prompt that makes LLMs align best with human reasoning is ``adopt a utilitarian/consequentialist framing'', but this does not mean that human participants explicitly considered consequentialist principles when providing judgments for the different scenarios. Indeed, there are scenarios where human participants make judgments that go against utilitarian principles. For example, in one of the stories, participants are asked whether it's morally permissible to kill 5 children of a particular race in an orphanage in order to save hundreds of other children in WWII. Here, participants overwhelmingly refuse to kill even if it means that this would have resulted in the greatest number of lives saved. So, as the reviewer suggests, while utilitarian principles certainly seem to play an important part in human moral reasoning, it's not the only consideration that matters to humans.

Different people's moral intuitions vary substantially. For some individuals, utilitarian principles matter a lot, whereas, for other individuals, deontological rules that prescribe what actions are right or wrong (e.g., do not kill) have a strong impact on their judgments. Currently, our evaluation of LLM alignment is against the aggregate of human judgments, and here it looks like prompting the LLM to adopt a utilitarian framing leads to the best alignment. That said, it's possible that different prompts would help the model to align better with different subgroups of human participants. While we are ultimately interested in better capturing the variance in human moral and causal intuitions, we chose not to focus on this aspect and leave it for future work.

\subsection{Additional Details on Average Marginal Component Effect Experiment}
\label{sec:app:amce}

\begin{figure}[h]
  \centering
  \includegraphics[width=\linewidth]{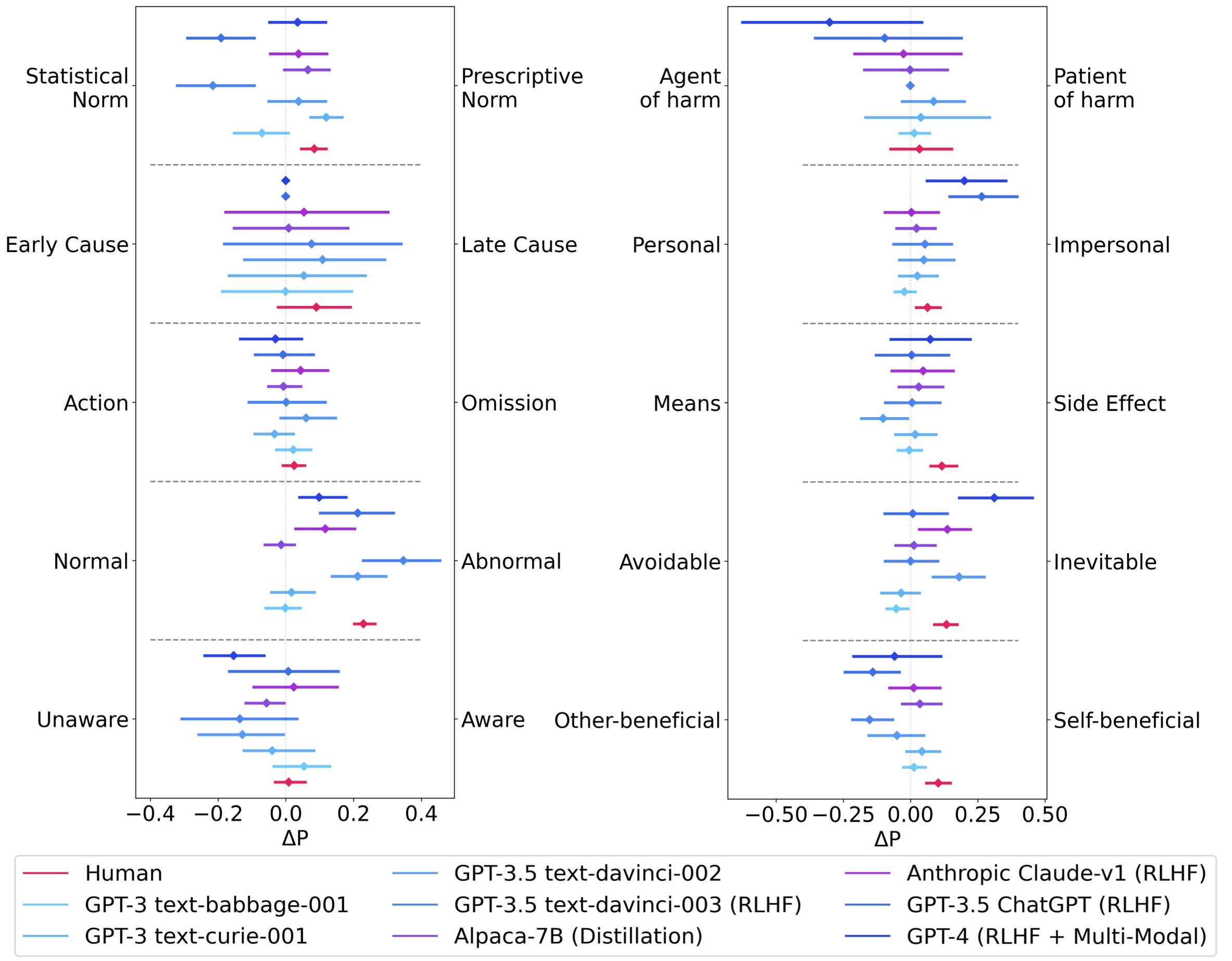}
  \caption{\textbf{AMCE for Different LLMs}: We show the 95\% bootstrapped confidence interval on AMCE.}
  \label{fig:factor_distribution}
\end{figure}

\begin{figure}[h]
  \centering
  \includegraphics[width=\linewidth]{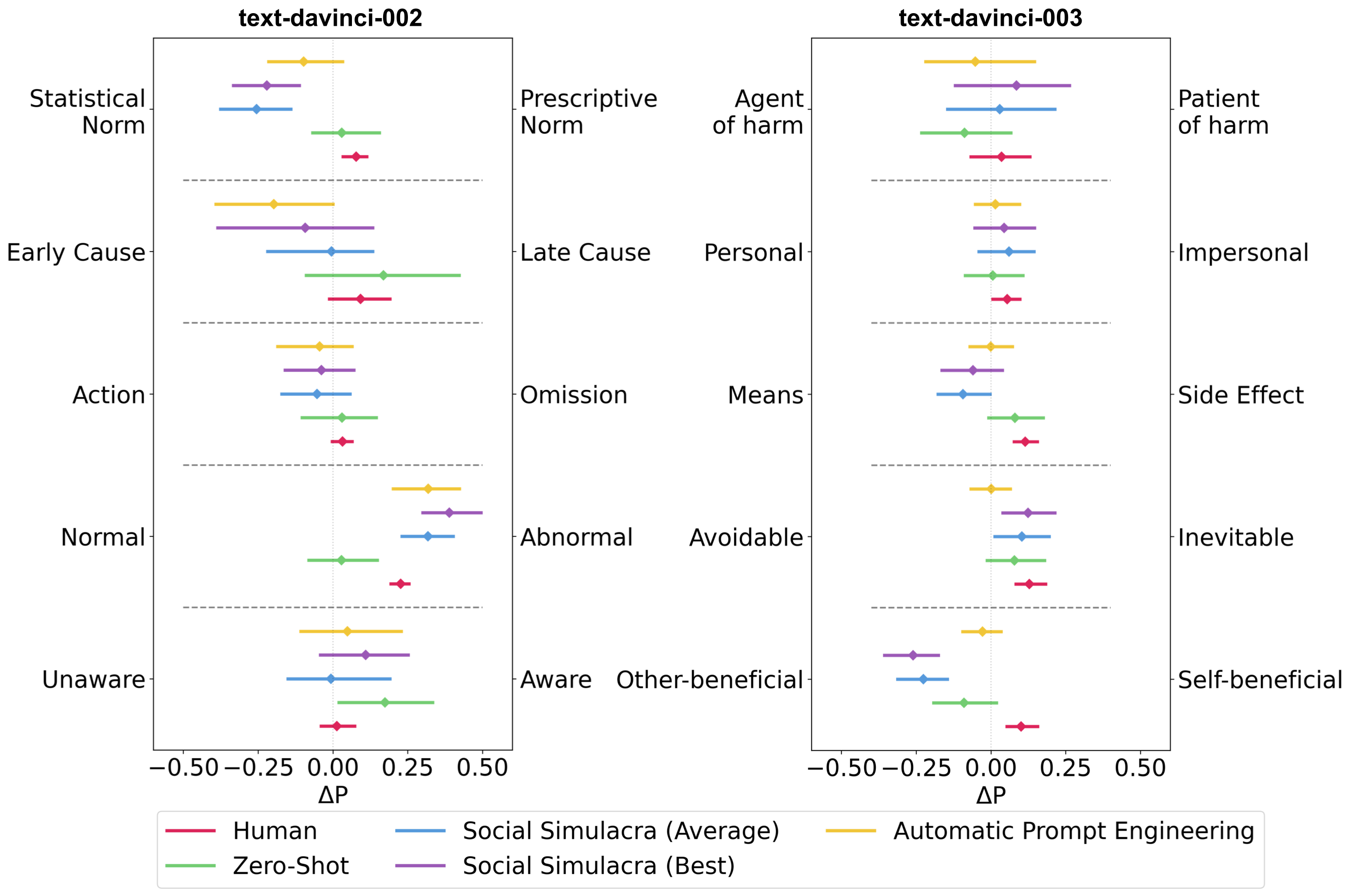}
  \caption{\textbf{AMCE for LLM with Prompting Methods}: We show that, with different prompting methods, the AMCE could change.}
  \label{fig:method_factor_distribution}
\end{figure}

We discuss the three assumptions that the conjoint analysis needs to have in order to claim the non-parametric estimator we used is an unbiased estimation of the AMCE. The first assumption is the stability and no carryover effect. In our context, this means the set of factor attributes from the story fully determines how the respondents will produce their responses. Since each of our stories has been used in previous scientific experiments and has validated scientific hypotheses, we believe these stories were written very carefully to highlight the factor attributes within the story.
The second assumption is no partial order effect. All the stories are independent, and we randomly sample the stories to present to the respondents. The second assumption is thus satisfied.
The third and last assumption is the randomization of factors. This assumption is satisfied in \cite{awad2018moral} because the scenarios in their experiment are randomly generated. We argue by collecting stories from a large number of papers, and with each paper focusing on investigating different scientific theories, the factors present in each story should be random if they are drawn from different studies. However, we acknowledge that because we are performing a meta-analysis over existing experiment designs, the stories we collected might be implicitly biased by our selection process. So it is possible that the AMCE we compute is a biased estimate. We compute confidence intervals via bootstrapping on AMCE and report them in \autoref{fig:factor_distribution} and \autoref{fig:method_factor_distribution}. Any ACME effect that crosses over 0.0 should be regarded as our analysis is inconclusive about whether the given model or human has a clear tendency given our dataset.

\subsection{Annotation Guidelines}

We provide the annotation guidelines as PDFs in our supplementary material. The annotator has a master's degree, and the annotation guidelines were developed jointly by researchers (authors) affiliated with computer science and psychology.

\subsection{Additional Statistics on Data}

\begin{figure}[h]
\centering
\begin{subfigure}{.5\textwidth}
  \centering
  \includegraphics[width=\linewidth]{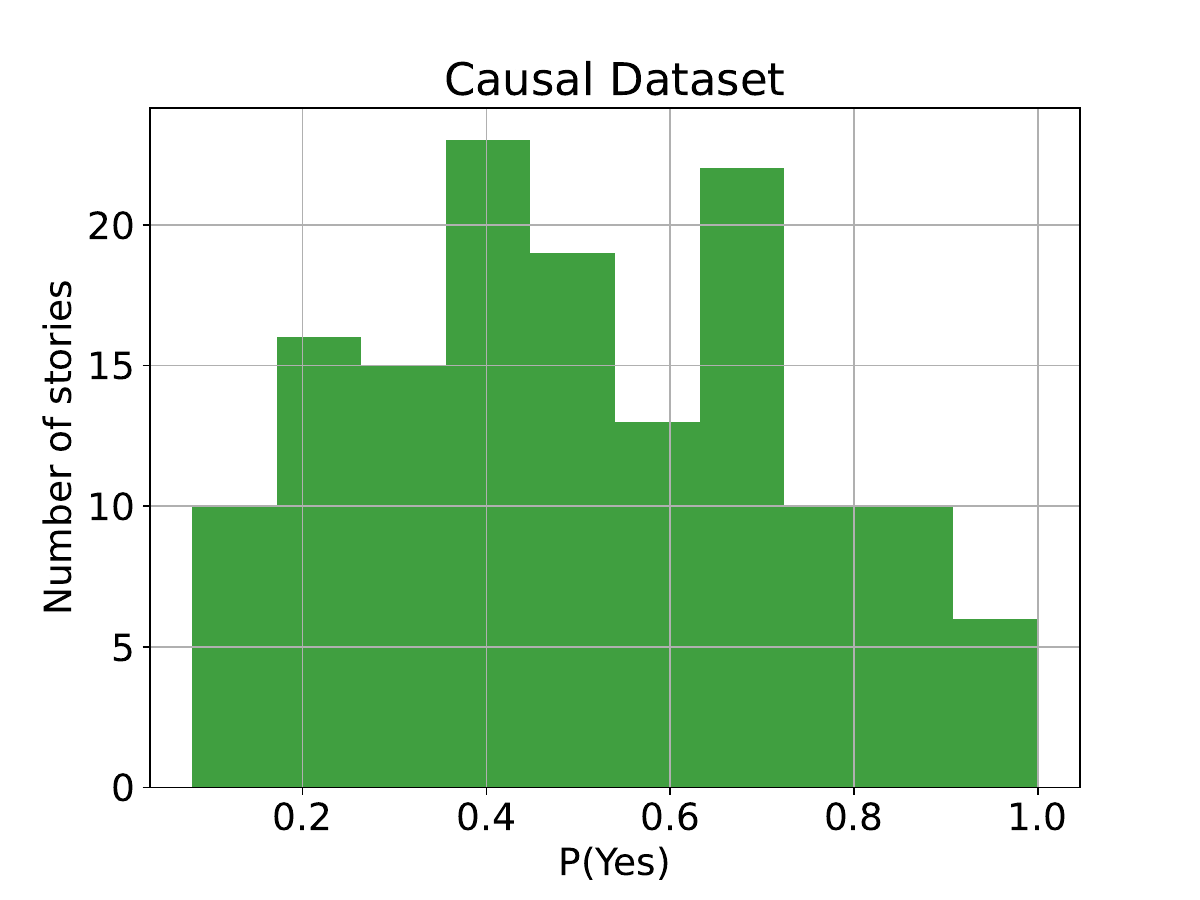}
  \caption{}
  \label{fig:sub1}
\end{subfigure}%
\begin{subfigure}{.5\textwidth}
  \centering
  \includegraphics[width=\linewidth]{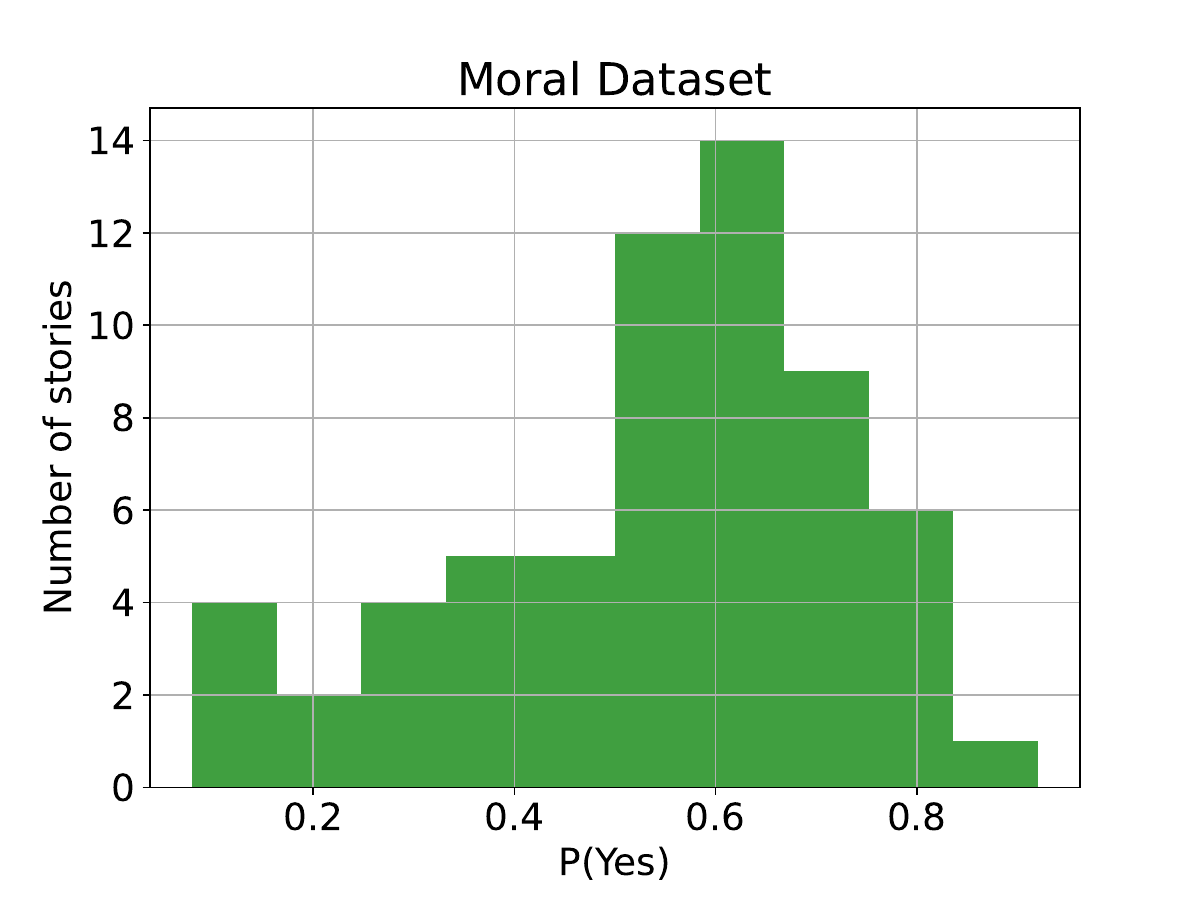}
  \caption{}
  \label{fig:sub2}
\end{subfigure}
\caption{The probability distribution of P(Yes) for aggregate human responses of Yes and No over 25 participants for each story.}
\label{fig:p_yes}
\end{figure}

We provide two figures on the distribution of P(Yes) for the stories in our dataset. This corroborates our claim that human causal and moral reasoning are diverse. We show the distribution in \autoref{fig:p_yes}.

\subsection{Factor Inference Experiment}
\label{sec:app:factor_infer}

We test if LLMs like GPT-3 have the ability to infer and assign the correct tag for the text segments in the story. We manually write a few examples of high-level instructions for each of the factors in \autoref{tab:causal_tag_defs}. We include all the examples we used in the appendix and provided in our code.

We find that LLMs such as GPT-3 are very good at inferring tags based on a few examples -- especially for factors that can be clearly specified by instructions and examples. We report the performance of inferring causal and moral factors in the story in \autoref{tab:causal_structure_result} and \autoref{tab:moral_structure_result}. Though GPT-3 performed poorly in the original story, on the causal factor inference task, the lowest accuracy is 74.7\%, and the highest is 97.3\%. This means that even though GPT-3 is unable to directly generate the human-matching answer for causal stories -- they can accurately recognize and assign the right tags for the text segments with minimal guidance from an expert human.

However, there are also factors that are more difficult to clearly define for LLMs. Generally, GPT-3 performed much worse on the moral structural inference task (\autoref{tab:moral_structure_result}), with a performance from 57.9\% to 86.0\%. For example, what ``causal role'' a person's action played in a scenario is a particularly difficult category for GPT-3 to correctly annotate.

\begin{table}[t]
\caption{\textbf{Factor Inference Experiment}: We test if GPT-3 models can correctly analyze a sub-segment of stories (extracted by human annotators). We run models 5 times to compute an average score and 95\% confidence intervals.}
\begin{subtable}[t]{\textwidth}
\caption{\textbf{Causal Factor Inference}}
\label{tab:causal_structure_result}
\small
\centering
\resizebox{\columnwidth}{!}{\begin{tabular}{@{}rccccccc@{}}
\toprule
GPT3 Size           & \begin{tabular}[c]{@{}c@{}}Weighted\\Average\end{tabular}        & \begin{tabular}[c]{@{}c@{}}Action \\ Omission\end{tabular} & Norm Type      & Time           & \begin{tabular}[c]{@{}c@{}}Agent \\ Awareness\end{tabular} & \begin{tabular}[c]{@{}c@{}}Causal \\ Structure\end{tabular} & \begin{tabular}[c]{@{}c@{}}Event \\ Normality\end{tabular} \\ 
\multicolumn{1}{l}{} & (N=497)        & (N=112)                                                    & (N=88)         & (N=67)         & (N=43)                                                     & (N=98)                                                      & (N=89)                                                     \\ \midrule
Majority       & 68.2            & 71.4                                                       & 76.1           & 86.6           & 53.5                                                       & 67.3                                                        & 50.6                                                       \\ \midrule
GPT3-babbage-v1           & 64.7 $\pm$ 3.5 & 80.5 $\pm$ 2.4                                             & 68.2 $\pm$ 3.7 & 35.8 $\pm$ 7.1 & 88.8 $\pm$ 3.2                                             & 58.6 $\pm$ 7.3                                              & 58.0 $\pm$ 8.2                                             \\
GPT3-curie-v1      & 82.8 $\pm$ 0.8 & 88.8 $\pm$ 2.3                                             & 91.4 $\pm$ 1.3 & 81.8 $\pm$ 2.0 & 96.7 $\pm$ 4.8                                             & 76.1 $\pm$ 4.1                                              & 68.1 $\pm$ 2.9                                             \\
GPT3.5-davinci-v2      &\textbf{85.5 $\pm$ 1.0} & 88.0 $\pm$ 2.2                                             & \textbf{97.3 $\pm$ 0.8} & 81.5 $\pm$ 2.1 & \textbf{97.2 $\pm$ 2.4}                                             & 74.7 $\pm$ 3.4                                              & \textbf{80.0 $\pm$ 1.8}                                             \\ \bottomrule
\end{tabular}}
\end{subtable}

\begin{subtable}[t]{\textwidth}
\caption{\textbf{Moral Factor Inference}}
\label{tab:moral_structure_result}
\small
\centering
\resizebox{\columnwidth}{!}{\begin{tabular}{@{}rcccccc@{}}
\toprule
GPT3 Size  & \begin{tabular}[c]{@{}c@{}}Weighted\\ Average\\ (N=202)\end{tabular} & \begin{tabular}[c]{@{}c@{}}Personal \\ Force\\ (N=48)\end{tabular} & \begin{tabular}[c]{@{}c@{}}Beneficience\\ (N=48)\end{tabular} & \begin{tabular}[c]{@{}c@{}}Locus of \\ Intervention\\ (N=48)\end{tabular} & \begin{tabular}[c]{@{}c@{}}Causal Role\\ (N=48)\end{tabular} & \begin{tabular}[c]{@{}c@{}}Counterfactual\\ Evitability\\ (N=48)\end{tabular} \\ \midrule
Majority   & 55.4 & 50.0 & 54.2 & 60.0 & 62.5 & 54.2                                                                        \\ \midrule
GPT3-babbage-v1       & 51.9 $\pm$ 2.5                                                       & 56.7 $\pm$ 5.6                                                     & 50.8 $\pm$ 5.0                                                & 60.0 $\pm$ 15.2                                                           & 40.0 $\pm$ 7.2                                               & 58.3 $\pm$ 5.2                                                                \\
GPT3-curie-v1 & 65.2 $\pm$ 2.7                                                       & 67.5 $\pm$ 5.4                                                     & 53.7 $\pm$ 1.2                                                & 66.0 $\pm$ 14.2                                                           & 59.2 $\pm$ 7.0                                               & \textbf{80.4 $\pm$ 2.9}                                                                \\
GPT3.5-davinci-v2 & \textbf{70.9 $\pm$ 2.4}                                                       & \textbf{72.9 $\pm$ 7.3}                                                     & \textbf{77.5 $\pm$ 2.2}                                                & \textbf{86.0 $\pm$ 14.2}                                                          & 57.9 $\pm$ 5.0                                               & 72.1 $\pm$ 5.4                                                                \\ \bottomrule
\end{tabular}}
\end{subtable}
\end{table}

We report the weighted recall and precision for the factor attribute inference task. We show that with limited instructions, GPT3.5-davinci-v2 can infer the correct attributes with high precision and recall for the causal factor inference task.

\begin{table}[h]
\caption{\textbf{Factor Inference Experiment (Recall)}:We report weighted recall for each of the subtask. We \textbf{boldface} results that outperform the majority baseline.}
\label{tab:r3_causal_moral_factor_recall}
\centering
\begin{subtable}[t]{0.85\textwidth}
\caption{\textbf{Causal Factor Inference (Recall)}}
\centering
\resizebox{\columnwidth}{!}{\begin{tabular}{@{}rcccccc@{}}
\toprule
GPT3 Size         & \begin{tabular}[c]{@{}c@{}}Action \\ Omission\end{tabular} & Norm Type      & Time           & \begin{tabular}[c]{@{}c@{}}Agent \\ Awareness\end{tabular} & \begin{tabular}[c]{@{}c@{}}Causal \\ Structure\end{tabular} & \begin{tabular}[c]{@{}c@{}}Event \\ Normality\end{tabular} \\ 
\multicolumn{1}{l}{}         & (N=112)                                                    & (N=88)         & (N=67)         & (N=43)                                                     & (N=98)                                                      & (N=89)                                                     \\ \midrule
GPT3-babbage-v1                     & 28.6            & 76.1      & 19.4 & 69.8            & 32.7             & 49.4            \\
GPT3-curie-v1                     & 32.1            & 23.9      & 7.5  & 46.5            & 67.3             & 49.4            \\
GPT3.5-davinci-v2               & 92.0            & 97.7      & 85.1 & 100.0           & 98.0             & 84.3            \\ \bottomrule
\end{tabular}
}
\end{subtable}

\vspace{0.2cm}

\centering
\begin{subtable}[t]{0.85\textwidth}
\caption{\textbf{Moral Factor Inference (Recall)}}
\label{tab:moral_factor_recall}
\small
\centering
\resizebox{\columnwidth}{!}{\begin{tabular}{@{}rccccc@{}}
\toprule
GPT3 Size & \begin{tabular}[c]{@{}c@{}}Personal \\ Force\\ (N=48)\end{tabular} & \begin{tabular}[c]{@{}c@{}}Beneficience\\ (N=48)\end{tabular} & \begin{tabular}[c]{@{}c@{}}Locus of \\ Intervention\\ (N=48)\end{tabular} & \begin{tabular}[c]{@{}c@{}}Causal Role\\ (N=48)\end{tabular} & \begin{tabular}[c]{@{}c@{}}Counterfactual\\ Evitability\\ (N=48)\end{tabular} \\ \midrule
GPT3-babbage-v1 & 50.0           & 54.2         & 40.0                  & 37.5        & 56.2                       \\
GPT3-curie-v1   & 50.0           & 50.0         & 40.0                  & 62.5        & 54.2                       \\
GPT3.5-davinci-v2  & 75.0           & 75.0         & 90.0                  & 64.6        & 68.8                       \\ \bottomrule
\end{tabular}
}
\end{subtable}
\end{table}

\begin{table}[h]
\caption{\textbf{Factor Inference Tasks (Precision)}: We report weighted precision for each of subtask. We \textbf{boldface} results that outperform the majority baseline.}
\label{tab:r3_causal_moral_factor_precision}
\centering
\begin{subtable}[t]{0.85\textwidth}
\caption{\textbf{Causal Factor Inference (Precision)}}
\centering
\resizebox{\columnwidth}{!}{\begin{tabular}{@{}rcccccc@{}}
\toprule
GPT3 Size         & \begin{tabular}[c]{@{}c@{}}Action \\ Omission\end{tabular} & Norm Type      & Time           & \begin{tabular}[c]{@{}c@{}}Agent \\ Awareness\end{tabular} & \begin{tabular}[c]{@{}c@{}}Causal \\ Structure\end{tabular} & \begin{tabular}[c]{@{}c@{}}Event \\ Normality\end{tabular} \\ 
\multicolumn{1}{l}{}         & (N=112)                                                    & (N=88)         & (N=67)         & (N=43)                                                     & (N=98)                                                      & (N=89)                                                     \\ \midrule
GPT3-babbage-v1 & 8.2             & 58.0      & 72.6 & 76.0            & 10.7             & 24.4            \\
GPT3-curie-v1   & 79.9            & 5.7       & 0.6  & 21.6            & 62.2             & 24.4            \\
GPT3.5-davinci-v2 & 92.8            & 97.8      & 86.4 & 100.0           & 98.0             & 86.1            \\ \bottomrule
\end{tabular}
}
\end{subtable}

\vspace{0.2cm}

\centering
\begin{subtable}[t]{0.85\textwidth}
\caption{\textbf{Moral Factor Inference (Precision)}}
\label{tab:moral_factor_precision}
\small
\centering
\resizebox{\columnwidth}{!}{\begin{tabular}{@{}rccccc@{}}
\toprule
GPT3 Size & \begin{tabular}[c]{@{}c@{}}Personal \\ Force\\ (N=48)\end{tabular} & \begin{tabular}[c]{@{}c@{}}Beneficience\\ (N=48)\end{tabular} & \begin{tabular}[c]{@{}c@{}}Locus of \\ Intervention\\ (N=48)\end{tabular} & \begin{tabular}[c]{@{}c@{}}Causal Role\\ (N=48)\end{tabular} & \begin{tabular}[c]{@{}c@{}}Counterfactual\\ Evitability\\ (N=48)\end{tabular} \\ \midrule
GPT3-babbage-v1 & 25.0           & 29.3         & 16.0                  & 14.1        & 57.7                       \\
GPT3-curie-v1   & 25.0           & 28.3         & 16.0                  & 39.1        & 29.3                       \\
GPT3.5-davinci-v2 & 78.1           & 77.5         & 91.4                  & 65.3        & 81.4                       \\ \bottomrule
\end{tabular}
}
\end{subtable}
\end{table}

\subsection{Prompts for Factor Inference}
\label{sec:app:tag-inference}

We also designed the prompts for asking GPT-3 to infer factor attributes. Unlike the standard setup in few-shot LLM inference tasks, where K samples are randomly chosen from a pool of training data, we manually write high-level instructions for GPT-3 or makeup examples that are different from the data. Also, unlike randomly sampling K examples, we treat our manually written prompt as ``instructions'' and keep it the same across examples. Although this strategy is very simplistic, we find it to work well for our task. 

We provide the full set of prompts in the supplement. Here are some examples. For causal factor inference, we show an example of inferring conjunctive or disjunctive causal structure. Note how high-level our description is. 

\begin{displayquote}
Both A and B need to happen in order for C to happen.

Is this conjunctive or disjunctive? 

Answer: Conjunctive.
\newline

Either A or B (or both) need to happen in order for C to happen.

Is this conjunctive or disjunctive? 

Answer: Disjunctive.
\newline

\texttt{[Text Segment to Infer]}

Is this conjunctive or disjunctive? 

Answer: \texttt{[LLM Generated Answer]}
\end{displayquote}

For moral factor inference, we show an example of inferring personal and impersonal force:

\begin{displayquote}

In the scenario, I pushed a button to trap another person.

Is the action personal or impersonal?

Answer: Impersonal.
\newline

In the scenario, my action directly affected another person through the use of physical force.

Is the action personal or impersonal?

Answer: Personal.
\newline

\texttt{[Text Segment to Infer]}

Is the action personal or impersonal?

Answer: \texttt{[LLM Generated Answer]}
\end{displayquote}

\subsection{Disambiguation: Factor vs. Factor-Attributes}

In the paper, we use ``factor'' to refer to high-level concepts that can be attributed to a text segment. For example, ``location'' is a higher-level concept and maps to ``factor'' in our definition. However, the underlying specific location, such as ``hotel'' or ``restaurant'' -- we address as ``attributes''. It is crucial to know that they correspond to two separate steps: the model should identify the factor for which the text segment contains information. Then, the model should identify the underlying attribute for this factor. This is a rather complicated process, and in this paper, we focus on 
the second step -- given a pre-selected text segment with a known factor for that segment (by a human), whether LLMs can infer the underlying attributes accurately. We find that LLMs can accomplish this task relatively well. We leave the investigation of the full pipeline to future work.

\subsection{Delphi API Response Labeling Guideline}

Delphi has provided a FreeQA API where we can submit the text, and the returned response falls under three classes: \textit{good}, \textit{discretionary}, and \textit{bad}. Since we are calculating our accuracy using a 3-class strategy, we treat \textit{good} as ``Yes'', \textit{bad} as ``No'', and \textit{discretionary} as ``ambiguous''.

\subsection{Annotation Agreement Calculation}
\label{sec:app:inter-rator}

We calculate the agreement between two annotators on both causal and moral datasets. Each annotator is allowed to assign as many or as few factors per story as they choose. After assigning the factor, they are asked to select text inside the story that best supports their decision to assign the factor to the story.

We calculate the agreement by IoU (Intersection over Union). The factor-attribute assignment overlap between two annotators on the causal dataset is 0.8920. The moral dataset's factor annotation is pre-specified by the authors of the original papers, therefore needing no manual annotation. For text segments for each factor, we only compute IoU if both annotators agree the factor is present. Under this condition, the IoU for causal text segments is 0.8145, and for moral text segments, it is 0.8423.

We include our calculation in \textsc{compute\_correlation.py} file.

\subsection{Crowd Sourced Voting Experiment Design and Interface}
\label{sec:app:crowd-source}

We recruit participants from Prolific to give responses to our stories. For each story, we elicit a yes/no answer from a single participant. We pay wages equivalent to \$12/hr, and we only recruit participants from the US and UK, with English as their first language. Each participant is presented with 5-6 stories either from the causal dataset or the moral dataset. We present the story and question and then ask them to choose yes or no. The interface is designed so that the participant is presented with one story at a time. Participants are allowed to change their decisions to any story before final submission. We keep track of how long they spend on each story. We do not collect personal, private information of any sort from the participants except their yes/no responses. All the raw data from our experiment is provided in the supplement.

\subsection{Model API Card}

Since our works use many models that can only be accessed through API, we provide details of these models and access time in Table~\ref{tab:model_card}. We obtained early access to Anthropic's claude-v1, a chat model. Anthropic has since discontinued the API access, and we cannot use the same API anymore.

\begin{table*}[htbp]
\centering
\small
\begin{tabular}{llp{0.7\linewidth}}
\toprule
Model & Access Date & Details \\
\midrule
GPT3-babbage-v1 & 2023-04-03 & GPT3-babbage-001 model that involves supervised fine-tuning on human-written demonstrations. \\
GPT3-curie-v1 &	2023-04-03 & GPT3-curie-001 model that involves supervised fine-tuning on human-written demonstrations. \\
GPT3.5-davinci-v2 &	2023-04-03 & GPT3.5-davinci-002 model that involves supervised fine-tuning on human-written demonstrations. Derived from code-davinci-002. \\
GPT3.5-davinci-v3 &	2023-04-03 & GPT3.5-davinci-003 model that involves reinforcement learning (PPO) with reward models. Derived from GPT3.5-davinci-002. \\
\midrule
Anthropic-claude-v1 & 2023-04-03 & A model trained using reinforcement learning from human feedback. \\
GPT3.5-turbo & 2023-10-20 & Sibling model of GPT3.5-davinci-003 is optimized for chat but works well for traditional completions tasks as well. Snapshot from 2023-06-13. \\
GPT4 & 2023-10-20 & GPT-4 is a large multimodal model (currently only accepting text inputs and emitting text outputs) that is optimized for chat but works well for traditional completions tasks. Snapshot of gpt-4 from 2023-06-13. \\
\bottomrule
\end{tabular}
\caption{List of models that are accessed through API in this work, including 6 OpenAI models and 1 Anthropic model.  
}
\label{tab:model_card}
\end{table*}

\subsection{FAQs}

\paragraph{Are there domain experts in moral psychology/philosophy involved in the design process of this paper?}

All the factors mentioned in Table 2 are taken directly from the papers that conducted the original experiments. A graduate student and a moral/causal psychology professor were involved in carefully discussing and designing these factors. A thorough literature review was conducted to make sure these factors are comprehensive enough.

\subsection{Data Sheet}

\subsubsection{Motivation}

\begin{itemize}[leftmargin=*]

\item \textbf{For what purpose was the dataset created?} The dataset was created to evaluate language models on causal selection problems and moral trolley problems. 

\item \textbf{Who created the dataset (e.g., which team, research group) and on behalf of which entity (e.g., company, institution, organization)?} The dataset is created by a multidisciplinary team at (Anonymous) University with researchers in the background of Computer Science, Statistics, Natural Language Processing, and Psychology.
  
\item \textbf{Who funded the creation of the dataset?} This project did not receive external funding.

\item \textbf{Any other comments?} N/A

\end{itemize}

\subsubsection{Composition}

\begin{itemize}[leftmargin=*]

\item \textbf{What do the instances that comprise the dataset
    represent (e.g., documents, photos, people, countries)?} The instances are all text descriptions of scenarios, not based on real events or people.

\item \textbf{Does the dataset contain all possible instances, or is it
    a sample (not necessarily random) of instances from a larger set?}
Yes, there is some filtering -- for example, we did not transcribe ALL story snippets that the causal papers we cited. This is for the purpose of label balancing. We try to select pairs of stories based on whether there is a significant change in human judgments that were verified by the paper's experiments. This helps make our dataset more balanced. For the moral dilemma dataset, we transcribed all the texts from each paper, resulting in a large label imbalance. 

\item \textbf{What data does each instance consist of?} Text.

\item \textbf{Is there a label or target associated with each
    instance?} Yes, for each instance, we have a binarized Yes/No label.
  
\item \textbf{Is any information missing from individual instances?}
No 

\item \textbf{Are relationships between individual instances made
    explicit (e.g., users' movie ratings, social network links)?} Multiple examples can come from the same paper, which is used to measure the same or similar factors (scientific hypotheses). 
  
\item \textbf{Are there recommended data splits (e.g., training,
    development/validation, testing)?} The entire dataset should only be used for testing.
  
\item \textbf{Are there any errors, sources of noise, or redundancies
    in the dataset?} No

\item \textbf{Is the dataset self-contained, or does it link to or
    otherwise rely on external resources (e.g., websites, tweets,
    other datasets)?} The dataset is self-contained.

\item \textbf{Does the dataset contain data that might be considered
    confidential (e.g., data that is protected by legal privilege or
    by doctor-patient confidentiality, data that includes the content
    of individuals' non-public communications)?} No

\item \textbf{Does the dataset contain data that, if viewed directly,
    might be offensive, insulting, threatening, or might otherwise
    cause anxiety?} No

\item \textbf{Does the dataset relate to people?} Yes, we collected human responses on each story.

\item \textbf{Does the dataset identify any subpopulations (e.g., by
    age, gender)?} No

\item \textbf{Is it possible to identify individuals (i.e., one or
    more natural persons), either directly or indirectly (i.e., in
    combination with other data) from the dataset?} No

\item \textbf{Does the dataset contain data that might be considered
    sensitive in any way (e.g., data that reveals racial or ethnic
    origins, sexual orientations, religious beliefs, political
    opinions or union memberships, or locations; financial or health
    data; biometric or genetic data; forms of government
    identification, such as social security numbers, criminal
    history)?} No

\end{itemize}

\subsubsection{Collection Process}

\begin{itemize}[leftmargin=*]

\item \textbf{How was the data associated with each instance
    acquired?} The data (text) were transcribed from research papers. Papers sometimes would display multiple segments of the text in a table. Therefore, we assemble the text according to the table and experimental settings. We do have to interpret the experimental results based on the paper's table/figures. When papers report an actual mean response score, we add them to the data file (comment section); when the paper doesn't report a mean response score but instead provides a bar plot of ratings, we try to read the height of the bar. Note that since we binarize the responses to ``Yes'' and ``No'' -- we only need to compare the bar height to the median height, which is relatively easy to do.

\item \textbf{What mechanisms or procedures were used to collect the
    data (e.g., hardware apparatus or sensor, manual human curation,
    software program, software API)?} 
    Data were collected by two graduate students. Follow-up factor annotation was conducted on TagTog.

\item \textbf{If the dataset is a sample from a larger set, what was
    the sampling strategy (e.g., deterministic, probabilistic with
    specific sampling probabilities)?} N/A

\item \textbf{Who was involved in the data collection process (e.g.,
    students, crowd workers, contractors) and how they were compensated
    (e.g., how much were crowd workers paid)?}
    Two graduate students are the data collectors. One of them is the lead author of the paper, who does not seek compensation. The other one is compensated at \$15/hr. The data collection process took about 70 hours, and are compensated for \$1025 in total. The additional text span annotation took about 20 hours for \$25/hr and is compensated for \$500 in total. We additionally use Prolific to collect 25 human responses for each story for \$12/hr rate. In total, Prolific workers were compensated for \$2500 in total for 205 hours of annotation work.

\item \textbf{Over what timeframe was the data collected?} 
Does this
The original papers of the dataset were published between 1976 to 2021. We transcribed the data between March 2021 and August 2021. The text segment annotation happened between March 2022 and May 2022. The Prolific crowdsourcing annotation happened during the month of September 2022.

\item \textbf{Were any ethical review processes conducted (e.g., by an
    institutional review board)?} Yes, we obtained IRB approval for our study.

\item \textbf{Does the dataset relate to people?} Explained above.

\item \textbf{Did you collect the data from the individuals in
    question directly, or obtain it via third parties or other sources
    (e.g., websites)?} We accessed the papers through the paper publisher's websites. Our dataset does not include a copy of the original paper, except we do provide a comment that links each data to which paper it was transcribed from.

\item \textbf{Were the individuals in question notified about the data
    collection?} Yes. We have a consent and notification form at the beginning of every study.

\item \textbf{Did the individuals in question consent to the
    collection and use of their data?} Yes. We have a consent form at the beginning of every study.

\item \textbf{If consent was obtained, was the consenting individuals
    provided with a mechanism to revoke their consent in the future or
    for certain uses?} Yes, on the consent form, we provide the email, phone number, and address of the primary investigator (PI) for the participants to contact.

\item \textbf{Has an analysis of the potential impact of the dataset
    and its use on data subjects (e.g., a data protection impact
    analysis)been conducted?} All our participants' data have remained anonymous. We do not collect PPI (Private Personal Information).

\item \textbf{Any other comments?} N/A

\end{itemize}

\subsubsection{Preprocessing/cleaning/labeling}

\begin{itemize}[leftmargin=*]

\item \textbf{Was any preprocessing/cleaning/labeling of the data done
    (e.g., discretization or bucketing, tokenization, part-of-speech
    tagging, SIFT feature extraction, removal of instances, processing
    of missing values)?} We perform quality checks on our labels by analyzing how long each participant took to label the example. We rejected participant labels if they didn't spend an adequate amount of time. Additionally, for 16 stories we transcribed from ``Immoral Professors and Malfunctioning Tools: Counterfactual Relevance Accounts Explain the Effect of Norm Violations on Causal Selection (Kominsky, Phillips, 2019)'', we have to make them shorter because they are longer than some of the language models can handle. We removed some sentences that only add additional context to the story but do not contribute to overall causal judgment (i.e., sentences like ``Alex and Benni are very reliable and Tom is satisfied with their work.''). We still include the raw, unprocessed instances in our dataset. We did not modify any other instance.

\item \textbf{Was the ``raw'' data saved in addition to the preprocessed/cleaned/labeled data (e.g., to support unanticipated future uses)?} We add the original mean response score in the comment field of our data file whenever the original research paper provides the actual score. We also kept the original transcription text in the data file.

\item \textbf{Is the software used to preprocess/clean/label the instances available?} No

\item \textbf{Any other comments?} No

\end{itemize}

\subsubsection{Use}

\begin{itemize}[leftmargin=*]
\item \textbf{Has the dataset been used for any tasks already?} No

\item \textbf{Is there a repository that links to any or all papers or systems that use the dataset?} No. We hope future users of this dataset will cite this paper, and then all the follow-up papers/systems will show up in the Google Scholar search engine.

\item \textbf{What (other) tasks could the dataset be used for?} N/A

\item \textbf{Is there anything about the composition of the dataset or the way it was collected and preprocessed/cleaned/labeled that might impact future uses?} MoCa is not a certification task, i.e., if the language model achieves high performance on MoCa, it is human-aligned. MoCa is only an evaluation task that tests if the model’s behavior is similar to humans.  Our focus is narrow and only on certain aspects of alignment with humans. It cannot and should not be used to make a sweeping and general statement about AI-human alignment.

\item \textbf{Are there tasks for which the dataset should not be used?} N/A

\item \textbf{Any other comments?} No

\end{itemize}

\subsubsection{Distribution}

\begin{itemize}[leftmargin=*]

\item \textbf{Will the dataset be distributed to third parties outside of the entity (e.g., company, institution, organization) on behalf of which the dataset was created?} Yes. The link can be found in the paper.

\item \textbf{How will the dataset will be distributed (e.g., tarball on website, API, GitHub)?} Data will be stored in a GitHub repo.

\item \textbf{When will the dataset be distributed?} The data are available right now.

\item \textbf{Will the dataset be distributed under a copyright or other intellectual property (IP) license, and/or under applicable terms of use (ToU)?} The dataset is under a Creative Commons license (CC BY 4.0). 

\item \textbf{Have any third parties imposed IP-based or other restrictions on the data associated with the instances?} N/A

\item \textbf{Do any export controls or other regulatory restrictions apply to the dataset or to individual instances?} N/A

\item \textbf{Any other comments?} N/A

\end{itemize}

\subsubsection{Meta-Data for Stories}

We made detailed comments for each of our collected examples. We include the experiment conditions, which table/figure from the original paper we derive the true label, and if possible, we include the average Likert scale score from the human responses. We provide them as part of the appendix for ease of reading, but they are also accessible through the data files we provided in the supplementary material.

We only sub-sampled some stories to put here:

\begin{enumerate}[leftmargin=*,label={Story}]
\item Normality and actual causal strength (Dropbox) (Icard, Kominsky, Knobe, 2017). Experiment 1, Vignette 1. (Same as Kominsky et al. 2015). Condition: normative vs. abnormal, conjunctive. M=3.37 vs. M=5.61.
\item Normality and actual causal strength (Dropbox) (Icard, Kominsky, Knobe, 2017). Experiment 1, Vignette 1: Motion detector. Condition: normative vs. abnormal, disjunctive. M=3.25 vs. M=4.18. Prescriptive norm.
\item Normality and actual causal strength (Dropbox) (Icard, Kominsky, Knobe, 2017). Experiment 1, Vignette 3: Train. We skip "battery” because it's a moral Condition: normative vs. abnormal, conjunctive. Prescriptive norm.
\item Normality and actual causal strength (Dropbox) (Icard, Kominsky, Knobe, 2017). Experiment 1, Vignette 3: Train. We skip "battery” because it's a moral Condition: normative vs. abnormal, disjunctive. Prescriptive norm.
\item What you foresee isn't what you forget: No evidence for the influence of epistemic states on causal judgments for abnormal negligent behavior. (Murray, et al., 2021) (Dropbox). Experiment 1: Epistemic advantage is not necessary for abnormal inflation, Vignette 1 (No knowledge), 2 (Knowledge) x 2 (Normality). Normal vs. abnormal. (M = 7.50, SD = 1.50, n = 85)  vs. (M = 3.32, SD = 2.47, n = 77). Changed (Henne et al. 2017), add whether Kate noticed or not.
\item What you foresee isn't what you forget: No evidence for the influence of epistemic states on causal judgments for abnormal negligent behavior. (Murray, et al., 2021) (Dropbox). Experiment 1: Epistemic advantage is not necessary for abnormal inflation, Vignette 1 (No knowledge), 2 (Knowledge) x 2 (Normality). (M = 7.48, SD = 1.74, n = 85) vs. (M = 4.79, SD = 2.60, n = 101). Changed (Henne et al. 2017), add whether Kate noticed or not.
\item What you foresee isn't what you forget: No evidence for the influence of epistemic states on causal judgments for abnormal negligent behavior. (Murray, et al., 2021) (Dropbox). Experiment 2: Outcome expectation is not necessary for abnormal inflation, 2 (Knowledge) x 2 (Normality), Vignette 1 (No knowledge). (Mean = approx. 8) vs. (Mean = approx. 3).
\item What you foresee isn't what you forget: No evidence for the influence of epistemic states on causal judgments for abnormal 
\item Degrading causation  (O'Neill, et al., 2019). Experiment 2. Conjunctive Electricity, Number of Causes 3. Normal vs. Abnormal. We only picked out ones where we can binarize (>0.5). We modified the question from “To what extent did X cause Y" (“totally vs. “not at all") to “Did X cause Y" (“Yes" vs. “No"). Testing abnormal inflation.
\item Crossed Wires: Blaming Artifacts for Bad Outcomes (Sytsma, 2021), Study 1: Machine Case with Responsibility Attributions. (M=4.89, SD=1.94) vs. (M=2.83, SD= 1.86)
\item Crossed Wires: Blaming Artifacts for Bad Outcomes (Sytsma, 2021), Study 3: Machine Case with Movement. (M=5.03, SD=2.24) vs. (M=2.88, SD=2.07)
\item Causation, norm violation, and culpable control (Dropbox) (Alicke, et al., 2011), Study 2, condition 1. Cheat vs. Did not Cheat condition is not so interesting (moral violation is judged as more causal). However, in Did not Cheat condition, norm vs. counternorm is more nuanced. Norm vs. Counternorm. (M=4.12 vs. M=2.62) Figure 5.
\item Counterfactual thinking and recency effects in causal judgment (Dropbox) (Henne, et al., 2021). Experiment 1, Vignette 1 (Overdetermination, Early vs. Late). (M=35.33 vs. M=-3.44)
\item Counterfactual thinking and recency effects in causal judgment (Dropbox) (Henne, et al., 2021). Experiment 1, Vignette 1 
\item The good, the bad, and the timely: how temporal order and moral judgment influence causal selection, (Reuter, et al, 2014), Rule violation. Scenario 1 (no rule violation) vs. Scenario 10 (Zoe violates a rule). These results are more comparative because the raw experiment has 5 options: Alice, Zoe, Both, None of the two, Not sure.
\end{enumerate}

Annotations for moral stories (Trolley problems only). Moral Machine Dataset and Simplified Moral Machine Dataset are synthetic, we documented the rationale at the beginning of this appendix.

\begin{enumerate}[leftmargin=*,label={Story}]
\item Throwing a Bomb on a Person Versus Throwing a Person on a Bomb Intervention Myopia in Moral Intuitions (Dropbox) (Waldmann \& Dieterich, 2007), Experiment 2 (based on Experiment 1 Trolley version, but with modifications), agent intervention with harm to two people as a side effect (AI/S), (M = 4.85, SD = 1.01). AI/M (Agent intervention, use people as means) (M=4.74).
\item Inference of Intention and Permissibility in Moral Decision Making (Dropbox) (Kleiman-Weiner, et al., 2015). Trial 1 (2v1). Trolley dilemma story based on Mikhail, J. (2007). Universal moral grammar: Theory, evidence and the future. Trends in cognitive sciences, 11(4), 143–152. Trial 2 (1vB)
\item Inference of Intention and Permissibility in Moral Decision Making (Dropbox) (Kleiman-Weiner, et al., 2015). Trial 1 (2vB). Trolley dilemma story based on Mikhail, J. (2007). Universal moral grammar: Theory, evidence and the future. Trends in cognitive sciences, 11(4), 143–152. Trial 2 (Bv2)
\item Moral judgment reloaded: a moral dilemma validation study (Christensen, et al., 2014), Personal-Instrumental (1) vs. Impersonal-Accidental (2) -- BURNING BUILDING (a) vs. BURNING BUILDING (b) Mean = 3.3953488372093 vs. 5.27906976744186
\end{enumerate}